\documentclass[11pt,table]{article}
\pdfoutput=1

\usepackage[utf8]{inputenc} 
\usepackage[T1]{fontenc}    
\usepackage[in]{fullpage}
\usepackage{xurl}

\usepackage{booktabs}       
\usepackage{amsfonts}       
\usepackage{nicefrac}       
\usepackage{microtype}      
\usepackage{amsmath,amssymb,amsthm}
\usepackage{mathtools}
\usepackage{mathrsfs}
\usepackage{thmtools}
\usepackage{thm-restate}
\usepackage{comment}
\usepackage{parskip}
\usepackage{siunitx}
\usepackage{etoolbox}
\usepackage{enumitem}
\usepackage{graphicx}
\usepackage{subcaption}
\usepackage[textsize=scriptsize,textwidth=2.2cm]{todonotes}
\usepackage{tabularx}
\usepackage{xcolor} 
\usepackage[numbers,sort]{natbib}
\usepackage[colorlinks=true]{hyperref}
\usepackage{longtable}
\usepackage{afterpage}
\usepackage{etoc}

\usepackage{indentfirst}
\usepackage{multirow}

\newcolumntype{?}{!{\vrule width 1.3pt}}
\usepackage{makecell}

\usepackage{algorithm}
\usepackage{algorithmic}

\def\eg{\emph{e.g.~}}
\def\etal{{\em et al.~}}
\def\ie{\emph{i.e.~}}

\title{Empirical Upper Bound, Error Diagnosis and Invariance Analysis of Modern Object Detectors} 

\author{Ali Borji\footnote{Work done while at MarkableAI.} \\
aliborji@gmail.com}

\begin{document}

\date{}

\maketitle

\begin{abstract}
Object detection remains as one of the most notorious open problems in computer vision. Despite large strides in accuracy in recent years, modern object detectors have started to saturate on popular benchmarks raising the question of how far we can reach with deep learning tools and tricks. Here, by employing 2 state-of-the-art object detection benchmarks, and analyzing more than 15 models over 4 large scale datasets, we I) carefully determine the upper bound in AP, which is 91.6\% on VOC (test2007), 78.2\% on COCO (val2017), and 58.9\% on OpenImages V4 (validation), regardless of the IOU threshold. These numbers are much better than the mAP of the best model (47.9\% on VOC, and 46.9\% on COCO; IOUs=.5:.05:.95), II) characterize the sources of errors in object detectors, in a novel and intuitive way, and find that classification error (confusion with other classes and misses) explains the largest fraction of errors and weighs more than localization and duplicate errors, and III) analyze the invariance properties of models when surrounding context of an object is removed, when an object is placed in an incongruent background, and when images are blurred or flipped vertically. We find that models generate a lot of boxes on empty regions and that context is more important for detecting small objects than larger ones. Our work taps into the tight relationship between object detection and object recognition and offers insights for building better models. Our code is publicly available at \url{https://github.com/aliborji/Deetctionupper bound.git}.


\end{abstract}


\section{Introduction}

\begin{figure*}[t]
\begin{center}
   \includegraphics[width=\linewidth]{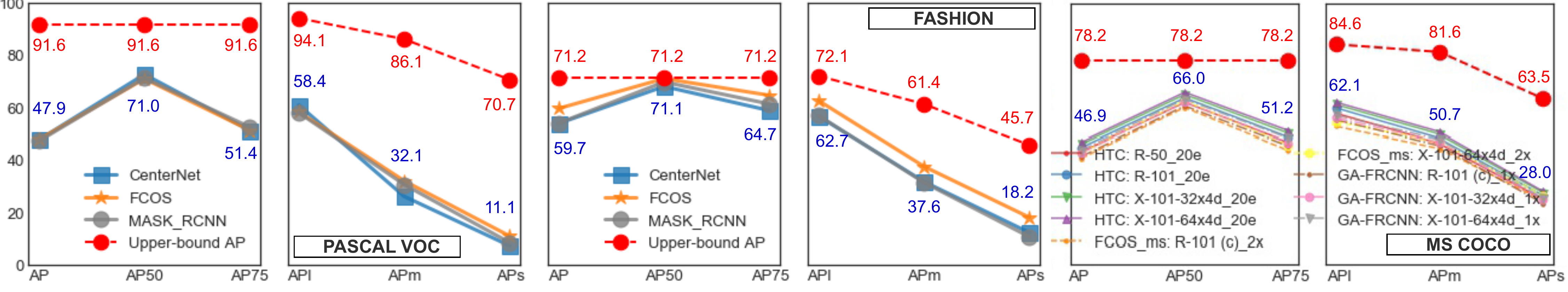}


   \caption{Upper bound AP (in red) and scores of the best model (in blue; FCOS~\cite{tian2019fcos} on VOC and FASHION, and Hybrid Task Cascade~\cite{chen2019hybrid} on COCO. Results show that scale remains the major problem in object detection.}

\label{fig:coco_fashion_voc}
\end{center}

\end{figure*}

Object recognition is believed to be (almost) solved in computer vision witnessed by the below human-error rate of state of the art models (about 3\% top-5 error on ImageNet~\cite{hu2018squeeze}) vs. about 5\% human error rate (although this number has not been carefully measured~\cite{russakovsky2015imagenet}). Object detection\footnote{The best published mAP (IOUs=.5:.95) on COCO2017 \underline{test-dev} is 51.0 by EfficientDet~\cite{tan2019efficientdet}. See \url{https://competitions.codalab.org/competitions/20794\#results} for the latest results on the COCO dataset.}, however, remains largely unsolved (66\% Avg. Prec. (AP) – even at 50\% overlap on COCOval2017; Fig.~\ref{fig:coco_fashion_voc}) which is far below the theoretical upper bound. Detection is much more challenging than recognition not only because precise localization is needed but because objects can undergo drastic transformations such as in-plane and in-depth rotation, scale, partial occlusions, etc. There is a larger variation of scale in detection datasets; the median scale of object instances relative to the image in ImageNet (classification) vs. COCO (detection) are 554 and 106, respectively. Therefore, most object instances in COCO are smaller than 1\% of the image area~\cite{singh2018analysis}. As such, detection can be considered as a litmus test for the capability of deep learning. 


Several years of extensive research on object detection\footnote{Please see~\cite{liu2018deep,zou2019object} for a review of generic object detection methods.} has resulted in accumulation of an overwhelming amount of knowledge regarding model backbones, tricks for model training and optimization, data collection and annotation, and model evaluation and comparison~\cite{zou2019object}, to a point that separating wheat from chaff is very difficult. As an example, truly understanding and implementing Avg. Prec. (AP) is frustratingly difficult. A quick Google search returns numerous blogs and codes with discrepant explanations of AP. To make matters even worse, it is not quite clear whether AP has started to saturate, whether progress is significant, and more importantly how far we can improve following the current path, making one wonder maybe we have reached the peak of performance using deep learning. Further, we do not know what is holding us back from making progress in object detection, compared to human-level (although debatable) accuracy of object recognition models. 

To shed light on the above matters, first we systematically and carefully approximate the empirical upper bound in AP. We hypothesize that the upper bound AP (UAP) is the score of the best recognition model that is trained on the training target bounding boxes and is then used to label the testing target boxes. We also investigate whether visual context surrounding a target object or its overlapping boxes can improve the upper bound AP. Second, we identify bottlenecks by characterising the type of errors that object detectors make and measure the impact of each one on performance. Third, we study the invariance properties of various object detectors on different types of transformations including incongruent context, scale, blur, vertical flip, etc.

In a nutshell, we find that there is a large gap between the performance of the best detection models and the empirical upper bound as shown in Fig.~\ref{fig:coco_fashion_voc}. This entails that there is a hope to reach this peak with the current tools, if we can find smarter ways to adopt object recognition models for object detection. We also find that classification remains as the major bottleneck in object detection and is more critical over small objects. Specifically, object detection models inherit the main limitations of CNNs which is the lack of invariance. Example failure cases include generating many bounding boxes on a white background containing a single object, and failing to detect objects in incongruent contexts, vertically flipped or blurred images. It seems that humans can still manage to solve these tasks, although with higher effort and lower performance than intact images.

\section{Related Work}

\subsection{Object recognition, semantic segmentation, and object detection: A unified view}
\label{unification}
\begin{figure}[t]
  \includegraphics[width=\linewidth]{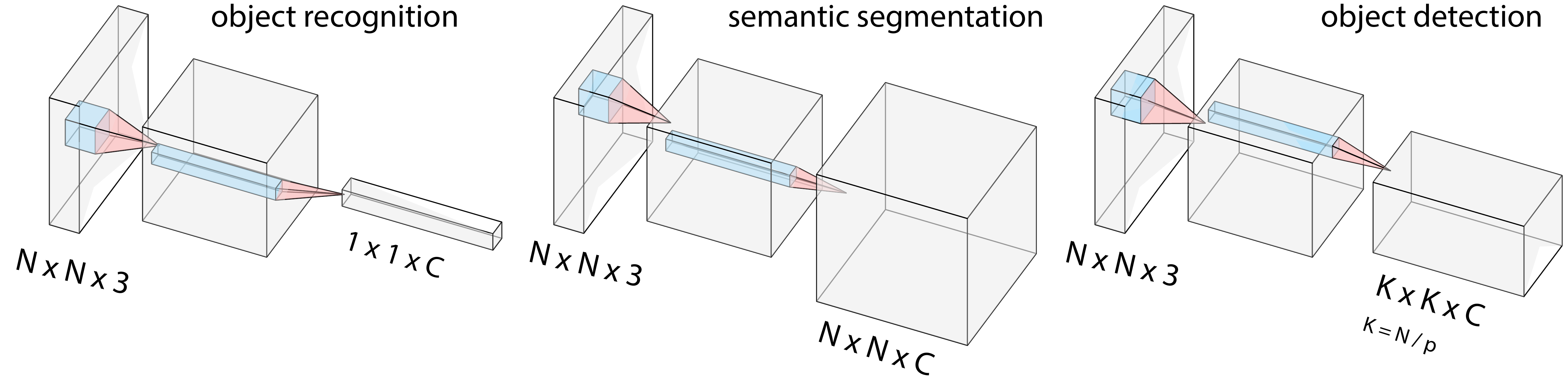}
  \caption{A unified view of object recognition, semantic segmentation, and object detection. $C$ is the number of classes.}
\label{fig:unify}
\end{figure}

A large number of architectures have been proposed in the past for three seemingly different tasks in computer vision namely object recognition, semantic segmentation and object detection. Here, we provide a unified view of these tasks, illustrated in Fig.~\ref{fig:unify}. In a simplified recognition architecture, a number of convloutional filters are applied to the input images (using a backbone) to generate a $C$ dimensional output vector where each element denotes the probability of the object belonging to a specific class. In semantic segmentation, the output consists of $C$ classes at the input image resolution. Each element denotes the probability of the image pixel belonging to a specific class (\eg sky, grass, car). Object detection falls somewhere in between to compromise speed vs. accuracy. For example, YOLO~\cite{redmon2016you} uses a grid as the output map ($C$ classes), where each cell contains information about few boxes/anchors at that location (\eg top-left position, width, height, objectness value). As another example, the output in CenterNet~\cite{zhou2019objects} consists of $C$ maps at the image resolution where activity at each pixel determines the probability of it being the center of an object. Additional maps are also used to predict width and height of the box centered at a point. As you can see, the resolution of output can be adjusted depending on whether we want to classify the entire image, every single pixel, or locate an object.



\subsection{Diagnosing object detection models}
Some of related works strive to understand detection approaches, identify their shortcomings, and pinpoint where more research is needed.
Parikh~\etal~\cite{parikh2011finding} aimed to find the weakest links in person detectors by replacing different components in a pipeline (\eg part detection, non-maxima-suppression) with human annotations. Mottaghi~\etal~\cite{mottaghi2015human} proposed human-machine CRFs for identifying bottlenecks in scene understanding. Hoeim~\etal~\cite{hoiem2012diagnosing} inspected detection models in terms of their localization errors, confusion with other classes, and confusion with the background on the PASCAL dataset. They also conducted a meta-analysis to measure the impact of object properties such as color, texture, and real-world size on detection performance. We replicate, simplify and extend this work on the larger COCO dataset and on image transformations.
Russakovsky~\etal~\cite{russakovsky2013detecting} analyzed the ImageNet localization task and emphasized on fine-grained recognition. Zhang~\etal~\cite{zhang2016far} measured how far we are from solving pedestrian detection. 
Vondrick~\etal~\cite{vondrick2013hoggles} proposed a method for visualizing object detection features to gain insights into their functioning. Some other related works in this line include~\cite{li2019analysis,zhu2012we,zhang2014predicting,goldman2019precise}. 


\subsection{Object detection benchmarks}
A number of studies strive to in compare object detection models. Some works have analyzed and reported statistics and performances over benchmark datasets such PASCAL VOC~\cite{everingham2010pascal,everingham2015pascal}, MSCOCO~\cite{lin2014microsoft}, CityScapes~\cite{cordts2016cityscapes}, and open images ~\cite{kuznetsova2018open}.
Recently, Huang~\etal~\cite{huang2017speed} performed a speed/accuracy trade-off analysis of modern object detectors. 
Dollar~\etal~\cite{dollar2011pedestrian} and Borji~\etal~\cite{borji2015salient,borji2012quantitative,borji2012state} compared models for person detection, and salient object detection, respectively. 
In \cite{michaelis2019benchmarking}, Michaelis~\etal assessed detection models on degraded images and observed about 30–60\% performance drop, which could be mitigated by data augmentation. In order to resolve the shortcomings of the AP score, some works have attempted to introduce alternative~\cite{hall2018probability} or complementary evaluation measures~\cite{oksuz2018localization,rezatofighi2019generalized}. A large number of works have also assessed object recognition models and their robustness (\eg~\cite{russakovsky2015imagenet,azulay2018deep,recht2019imagenet,mishkin2017systematic,borji2014human}).


\subsection{Contextual influences in object detection}
\begin{figure}[t]
\begin{center}
   \includegraphics[width=.7\linewidth]{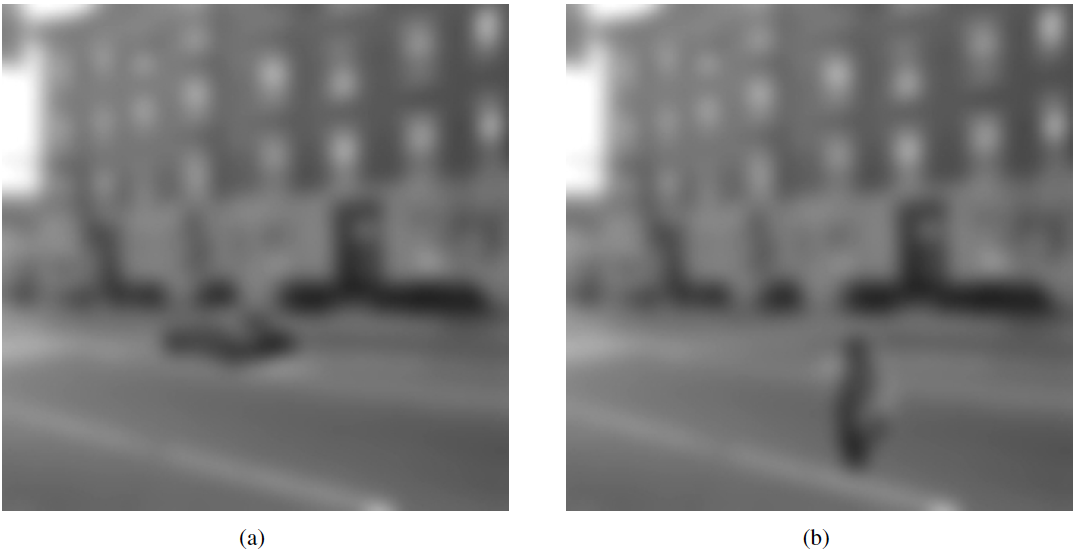}

   \caption{In presence of image degradation (\eg blur, noise), object recognition is heavily guided by contextual information. The object appearing as a car in (a) is rotated 90 degrees and is placed in image in panel (b) which now appears to be a pedestrian (image reproduced from~\cite{torralba2003contextual}).}
\vspace{-10pt}   
\end{center}
\label{fig:contexttorralba}
\end{figure}

Visual context is believed to be a rich source of information about an object’s identity,
location and scale, especially when appearance information is weak (See Fig.~\ref{fig:contexttorralba}). Torrabla \etal~\cite{torralba2003contextual} introduced a framework to model the relationship between
context and object properties based on the correlation between the statistics of low-level features across the entire
scene and it objects. Several other works have studied the role of context in object detection and recognition (\eg~\cite{bar2004visual,wolf2006critical,marat2012influence,heitz2008learning,torralba2001statistical,rabinovich2007objects,rosenfeld2018elephant,galleguillos2010context,zheng2009quantifying}. Heitz~\etal~\cite{heitz2008learning} proposed a probabilistic framework to capture contextual information between “stuff” and “things” to improve object detection on PASCAL VOC. Barnea~\etal~\cite{barnea2019exploring} utilized co-occurrence relations among objects to improve the detection scores. Divvala~\etal~\cite{divvala2009empirical} explored different types of context in recognition. See also~\cite{heitz2008learning,chen2018context,song2011contextualizing,hu2018gather,marat2012influence,alamri2019contextual,karianakis2016empirical,zhu2016object}. 


\section{Experimental Setup}

\subsection{Benchmarks}
We base our analysis on two recent large-scale object detection benchmarks: \emph{MMDetection}~\cite{chen2019mmdetection}\footnote{\url{https://github.com/open-mmlab/mmdetection}} and \emph{Detectron2}\footnote{\url{https://github.com/facebookresearch/detectron2}}. The former evaluates more than 25 models. The latter includes several variants of FastRCNN~\cite{girshick2015fast}. In both benchmarks, all COCO models have been trained on \emph{train2017} and evaluated on \emph{val2017}. Here, we use \emph{MMDetection} to train and test additional models on a new dataset.


\subsection{Models}
We consider the latest models published in the major vision conferences and the ones included in the above benchmarks. Several variants of the RCNN model including FasterRCNN~\cite{ren2015faster}, MaskRCNN~\cite{he2017mask}, RetinaNet~\cite{lin2017focal}, GridRCNN~\cite{lu2019grid}, LibraRCNN~\cite{pang2019libra}, CascadeRCNN~\cite{cai2018cascade}, MaskScoringRCNN~\cite{huang2019mask}, GAFasterRCNN~\cite{zhu2019empirical}, and Hybrid Task Cascade~\cite{chen2019hybrid} are considered. We also include SSD~\cite{liu2016ssd}, FCOS~\cite{tian2019fcos}, and CenterNet~\cite{zhou2019objects}. Different backbones for each model are also taken into account. 

\subsection{Datasets}
We employ 4 datasets including:

\begin{itemize}
    \item {\bf PASCAL VOC}~\cite{everingham2015pascal}: We use \emph{trainval0712} for training (16,551 images, 47,223 boxes) and \emph{test2007} (4,952 images, 14,976 boxes) for testing. This dataset has 20 categories. 
    \item {\bf FASHION dataset}: This dataset covers 40 categories of clothing items (39 + humans). Trainval, and test sets for this dataset contain 206,530 images (776,172 boxes) and 51,650 images (193,689 boxes), respectively.  Fig.~\ref{fig:fashion_stats1}.A displays samples from this dataset (and also additional statistics). This is a challenging dataset since clothing items are non-rigid as opposed to COCO or VOC objects.
    \item {\bf MS COCO}~\cite{lin2014microsoft}: MSCOCO has 80 categories. It has carried the torch for benchmarking advances in object detection for the past 6 years. We use \emph{train2017} for training (118,287 images, 860,001 boxes) and \emph{val2017} (5,000 images, 36,781 boxes) for testing.
    \item {\bf OpenImages}~\cite{kuznetsova2018open}: We use the OpenImages V4 dataset, used also in the Kaggle competition\footnote{\url{https://www.kaggle.com/c/open-images-2019-object-detection}}. It has 500 classes and contains 1,743,042 images (12,195,144 boxes) for training and 41,620 images (226,811 boxes) for validation (used here for testing).  
\end{itemize}

\afterpage{ 
\clearpage 
\begin{figure*}[htbp]
\begin{center}
\vspace{-50pt}
\hspace{-20pt}
   \includegraphics[width=1\linewidth]{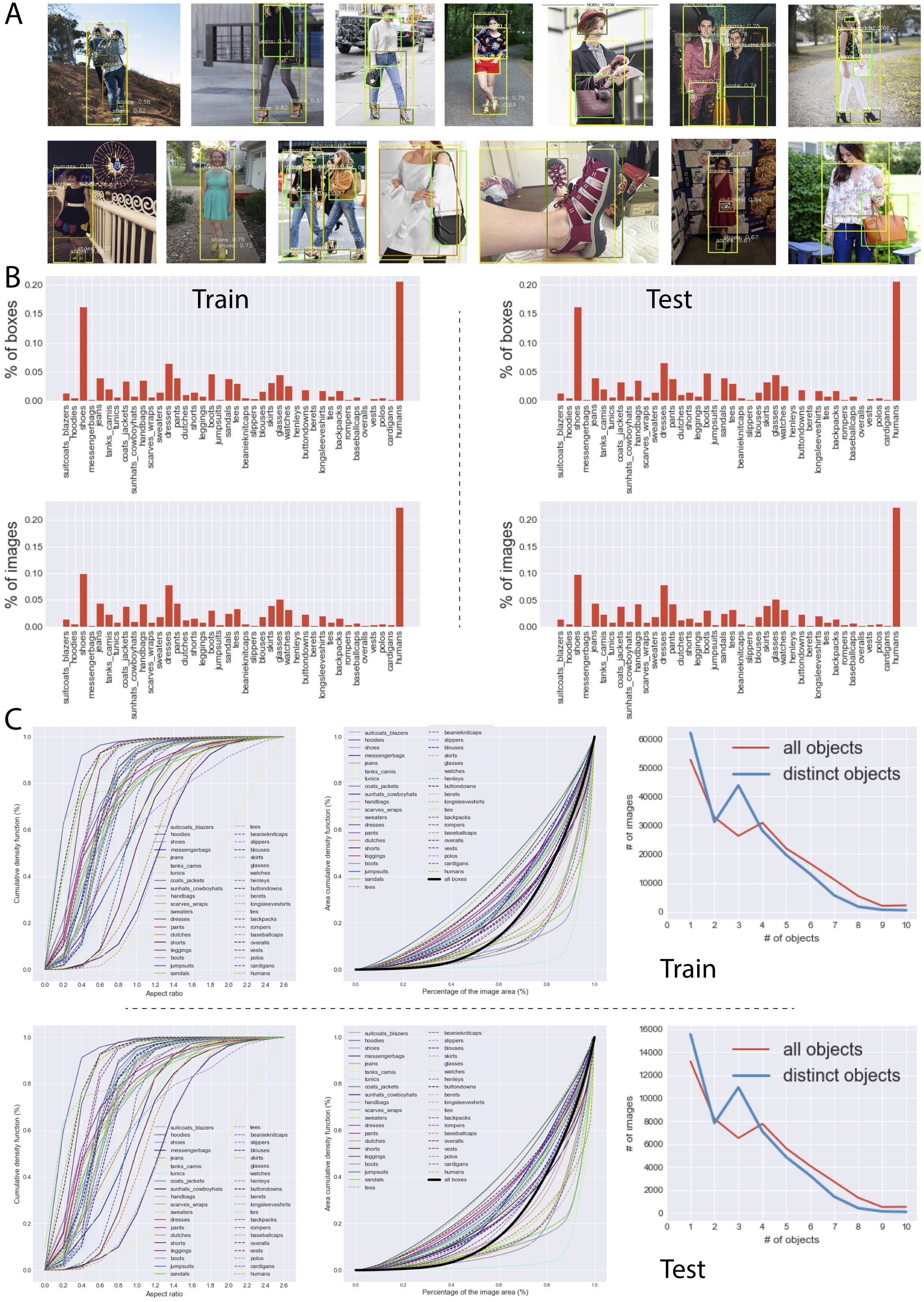}
\end{center}
\caption{Statistics of the FASHION dataset. A) Sample images along with FCOS predictions, B) Percentage of annotated bounding boxes and images in train and test sets, and C) Aspect ratio, object size, and number of objects per image.}
\label{fig:fashion_stats1}

\end{figure*}  

\thispagestyle{empty}
\clearpage 
}

\subsection{Metrics}
We use COCO evaluation code (\url{http://cocodataset.org/\#detection-eval}) to measure AP over IOU thresholds of 0.5 and 0.75 as well as the average AP over IOUs in the range 0.5:.05:0.95. APs are calculated per class and are then averaged. We also report breakdown APs over small (area$<32^2$), medium ($32^2<$area$<96^2$), and large (area$>96^2$) objects. 
See also \url{https://github.com/rafaelpadilla/Object-Detection-Metrics},~\url{http://cocodataset.org/\#detection-eval}, and~\url{https://medium.com/@kemal.oksz/which-one-to-measure-the-performance-of-object-detectors-ap-or-olrp-936d072a6eb0.}.




\section{Characterizing the Empirical Upper Bound}
\label{sec:uap}
We hypothesize that the empirical upper bound in AP is the score of a detector with ground truth bounding boxes labeled by the best object classifier. The classification score is considered as the detection score. This way we essentially assume that the localization problem is solved and what remains is only object recognition. 
However, it might be possible to improve upon this detector in at least two ways: a) by exploiting local context around an object to improve classification accuracy and hence better UAP, and b) by searching over the scene and finding boxes that are easier to classify (compared to the target box) and have enough overlap with the target box. This does not matter for the perfect IOU but may affect IOUs lower than one. 
We carefully investigate these possibilities in the following. 


\begin{figure}[t]
\begin{center}
  \includegraphics[width=\linewidth]{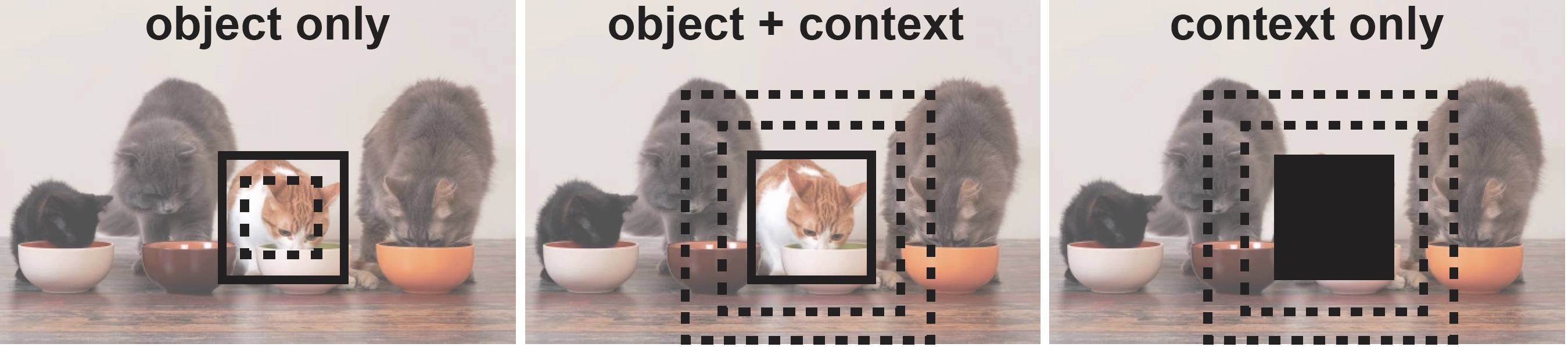}
\end{center}
  \caption{Illustration of visual context surrounding an object.}
\label{fig:context}
\end{figure}



\begin{table*}[htbp]
\begin{center}
\begin{small}
\renewcommand{\tabcolsep}{3.3pt}

\begin{tabular}{l|l|ccccc|ccccc|ccc}
  & \multirow{2}{*}{\textbf {Dataset}} & \multicolumn{5}{c|} {\textbf{object only}} & \multicolumn{5}{c|} {\textbf{object + context}} & \multicolumn{3}{c} {\textbf{context only}} \\
 \cline{3-15} 
test on & & 0.2 &  0.4 & 0.6 & 0.8 & 1 & 1.2 & 1.4 & 1.6 & 1.8 & 2 &      1.2 & 2 & all img  \\
\hline\hline
\multirow{4}{*}{\textbf{\rotatebox[origin=c]{90}{object}}} & VOC & 39.3 &	68.0	& 82.6&	92.5	& {\bf 94.8}&	93.0 &	91.6&	90.6&	88.6	& 87.0   & 63.6 &	64.9	& 35.3     \\
& FASHION & - &	52.9	& 66.4	&71.7	& {\bf 88.8}	&82.3&	77.2 &	71.8  & 67.9 &		64.8  &    29.0	& 32.2	&	12.0\\
& COCO & - & - & 67.1	 & 79.8 & 	{\bf 86.7} & 82.9	 & 78.3 & 	72.5	 & 67.4 & 	63.0    & 43.7 &	48.9	&	11.0\\
& OpenImages & - & - & -& - & {\bf 69.0} & 65.1 & 62.7 & - & -& - & - &  - & - \\
\hline \hline
\multirow{3}{*} {\textbf{\rotatebox[origin=c]{90}{+/- x}}} & VOC & 61.1 & 79 & 87.2 & 92.4 & {\bf 94.8} & 94.4 & 94.0 & 93.7 & 92.4 & 91.3    & 61.8 & 79.6 & 73.5 \\
& FASHION & - & 73.1 & 81.2 & 86.7 & {\bf 88.8} & 88.4 & 87.2 & 85.9 & 83.82 & 82.28 &   72.5 & 76.1 & 74.3\\
& COCO & - & - & 74 & 81.4 & 86.7 & 86.8 & 87.3 &  87.6 & {\bf 87.7} & 87.3  & 57.6 & 69.7 & 63.4 \\

\end{tabular}
\end{small}
\end{center}
\caption{Recognition accuracy using object and/or its context. Top rows: testing on the canonical object size (used in the rest of the paper). Bottom rows: training and testing are the same, for example, a classifier is trained on the object-only case 0.6 and is then tested on the object-only case 0.6. As you can see, results in this case are better. Bold font shows the maximum per row.}
\label{tab:context}
\end{table*}

\subsection{Utility of the surrounding context}
We trained ResNet152~\cite{he2016deep} on target boxes in three settings as shown in Fig.~\ref{fig:context}: I) \emph{object only}, II) \emph{object + context}, and III) \emph{context only}. Standard data augmentation techniques including mean pixel subtraction, color jittering, random horizontal flip and random rotation (10 degrees) were applied. Boxes were resized to 224 $\times$ 224 pixels and models were trained for 15 epochs. Trained models were tested on the original object box. Results (top-1 accuracy) shown in Table~\ref{tab:context} reveal that the canonical object size contains the most information regarding the category of an object over all four datasets\footnote{Please see Fig.~\ref{fig:CMs} for confusion matrices of these classifiers.}. Increasing or decreasing object box lowers the performance. Context-only scenario leads to high classification score but still does below other cases. Stretching the context to the whole scene drops the performance significantly. 
Training and testing models on the same condition (\ie both on \emph{object+context}) results in higher accuracy on that specific condition but does not lead to better overall recognition accuracy. 





\subsection{Searching for the best label}
Essentially the problem definition here is how we can get the best classification accuracy for recognition of objects in the scene by utilizing all the information in the scene. This is different than recognition approaches that treat objects in isolation. Note that recognition accuracy is not the same as AP, since detection scores also matter in AP calculation.

Having the best classifier at hand, we are ready to approximate the empirical upper bound in AP. Before delving into details first lets recap how AP is calculated.


\noindent {\bf AP calculation}. For each category, detections over all images are sorted according to their confidences. Starting from the top of this list, the target with the highest IOU with each detection is considered. We have a true positive (TP; hit) if their IOU is $\geq thresh$, and if that target has not been assigned yet. We have a false positive (FP) if IOU$<thresh$ (\ie localization error) or if the target has been assigned (\ie duplicate; two predictions on the same target). A target box can be matched with only one detection (the one with the highest confidence score and IOU$\geq thresh$). If a detection has IOU$\geq thresh$ with two targets, it is assigned to the one with the highest IOU which is not assigned already. Scanning the sorted detection list again, a precision for each recall is obtained and is used to draw the Recall-Precision (RP) curve and to compute the AP. 



\begin{figure}[t]
\begin{center}
   \includegraphics[width=\linewidth]{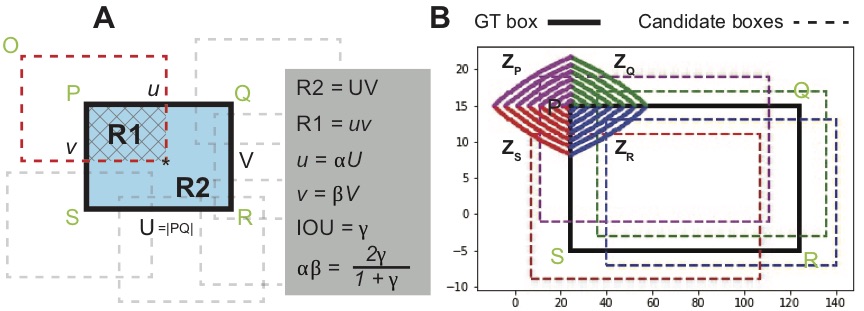}
\end{center}
   \caption{A) Illustration of our setup for finding boxes with IOU $\geq \gamma$ with the target box (corresponding to $\alpha\beta = 2\gamma/(1+\gamma)$; $\alpha\beta=2/3$ for $IOU=0.5$), B) The solutions are 4 curves represented by Eqs. 4 to 8. Four sample rectangles are shown with dashed lines.}
\label{fig:synthetic_rects}
\end{figure}

We explore two strategies in pursuit of the upper bound AP. In the {\bf first strategy}, we apply the best classifier from the previous section to the target boxes. The detector built in this fashion gives the same AP regardless of the IOU threshold, since our detections are target boxes. As we argued above, it is not possible to improve this detector at IOU=1. However, if we are interested in upper bound for a lower IOU (say $\gamma$), then it might be possible to do better by searching among the candidate boxes near a target box and choose the one that can be classified better than the target box, or aggregate information from nearby boxes. Thus, in our {\bf second strategy}, we sample boxes around an object and either apply the original classifier (trained on canonical object size) or train and test new classifiers on the surrounding boxes. In any case, we always keep the target box but change its label and/or its confidence. First, lets take a look at our box sampling approach, which is illustrated in Fig.~\ref{fig:synthetic_rects}.



\noindent {\bf Sampling boxes with IOU above a threshold.} 
Here, we are interested in finding the coordinates of the top-left corner of all rectangles with IOU $\geq \gamma \ \ \big(\gamma \leq 1\big)$ with the ground-truth bounding box. We use the coordinate system centered at the top-left corner of the target box $P$; which can be easily converted to the image level coordinate frame. The bottom-right coordinates of the desired rectangles that intersect with the target box from the top-left follow the equation $\alpha \beta = 2\gamma/(1 + \gamma)$, where $\alpha$ and $\beta$ are width and height of rectangles, respectively (we assume all boxes have the same width and height as the target box). According to the illustration in Fig.~\ref{fig:synthetic_rects}(A), we have:
\begin{equation}
\text{R1 = uv}  \text{, \ \  } \text{R2 = UV}  \text{, \ \  } \text{IOU} = \gamma  \text{, \ \  } \text{IOU} = \frac{R1}{2R2 - R1}   
\end{equation}
From these equations and assuming $u = \alpha U$, and $v = \beta V$, it is easy to derive the following equations:
\begin{equation}
    R1 = \alpha U \beta V \text{, \ \  } R1 = \frac{2\gamma}{1 + \gamma} R2  \\
\end{equation}
and also:
\begin{equation}
    \alpha \beta = \frac{2\gamma}{1 + \gamma} \text{, \ \ }  \alpha \beta = \frac{2}{3} \text{\ \ for \ \ }  \gamma = 0.5 
\end{equation}
The same equation governs the coordinates of the bottom-left, top-left, and top-right corners of the rectangles intersecting with the target box at points $Q$, $R$, and $S$, respectively (in the coordinate frames centered as these points, in order). Calculating the top-left corner of these rectangles (in their corresponding coordinate frames) and representing them in the coordinate frame of point $P$, we arrive at the following four equations (note that these are not lines): 
\begin{eqnarray}
Z_P : \ \ \big\langle~~ (\alpha - 1) U + x_P, \ \ (\beta - 1) V + y_P  ~~\big\rangle \\
Z_Q : \ \ \big\langle~~ (1 - \alpha) U + x_P, \ \ (\beta - 1) V + y_P ~~\big\rangle \\
Z_R : \ \ \big\langle~~ (1 - \alpha) U + x_P, \ \ (1 - \beta) V + y_P ~~\big\rangle \\
Z_S : \ \ \big\langle~~ (\alpha - 1) U + x_P, \ \ (1 - \beta) V + y_P ~~\big\rangle \\
\forall \text{ \ } \alpha,\beta \leq 1, \text{\ \ s.t. \ \ } \alpha \beta = \frac{2\gamma}{1 + \gamma}
\end{eqnarray}

\begin{table}[t]
\begin{center}
\renewcommand{\tabcolsep}{8pt}

\begin{tabular}{l|l|llll|llll} 


 \multirow{2}{*}{\textbf {Dataset}} & \multirow{2}{*}{\textbf {Acc.}} &  \multicolumn{4}{c|} {\textbf{Most Confident Box}} &  \multicolumn{4}{c} {\textbf{Most Frequent Label}} \\ 
 
\cline{3-10}

 & & $AP$ & $AP_{l}$ & $AP_{m}$ &  $AP_{s}$ & $AP$ & $AP_{l}$ & $AP_{m}$ &  $AP_{s}$  \\ 
 
\hline\hline
VOC & 93.7 & 88.7 & 91.7 & 81.4 & 63.8 &  89.1 & 92.0 & 82.9 & 60.0 \\
FASHION &   87.4 & 68.1 & 68.6 & \underline{61.9} & \underline{49.5}  & 67.7 & 68.2 & \underline{60.7} & \underline{47.8} \\
COCO & 84.8 & 76.9 & 81.8 & 80.6 & \underline{62.8} & 76.4 & 82.0 & 80.4 & 60.7 \\

\hline\hline


VOC &  88.5 & 91.3 & 82.5 & 61.8 & 89.4 & 90.3 & 92.9 & 84.7 & 63.3  \\
FASHION & 68.7  & 56.1 & 56.6 & 54.8 & 36.1 & 49.0 & 49.3 & 48.8 & 33.9\\
COCO & 74.8 & 73.0 & 77.6 & 76.7 & 60.3 & 70.4 & 73.1 & 74.6 & 58.2\\

\end{tabular}
\end{center}
\caption{Results of our second strategy for estimating the upper bound AP (\ie searching for the best bounding box or object label near a target box; among boxes with IOU $\geq 0.5$). Notice that upper bound for AP, AP$0.5$ and AP$0.75$ are all the same. Underlined numbers show where we could improve over the 1st strategy. Top rows) using a classifier trained on surrounding boxes, Bottom rows) using the original classifier trained on the canonical object size.}
\label{tab:control_res}
\end{table}

Using the above equations, we then sample some (here = 4) rectangles with $IOU\geq \gamma$ (Fig.~\ref{fig:synthetic_rects}(B)) and label them with the label of the target box. We then train a new classifier (same ResNet152 as above) on these boxes. This is effectively a new data augmentation technique. Notice that AP is a direct consequence of the classification accuracy, so if we can better classify objects we can obtain a better AP. To estimate UAP, we 
sample a number of rectangles (=4) near a target box (all with $IOU\geq \gamma$), and then label the target box with: a) the label (and confidence) of the box with the highest classification score (\ie most confident box), or b) the most frequent label among the nearby boxes (with the maximum confidence score among them). 

Just recently,~\cite{BBGenerator} proposed a similar solution to ours for generating bounding boxes. Their approach is more general and relaxes the constraint of boxes to have the same width and height. Please see Fig.~\ref{fig:kemal}.

\afterpage{ 
\clearpage 
\begin{figure*}[htbp]
\vspace{-70pt}

   \includegraphics[width=1.5\linewidth, angle=90]{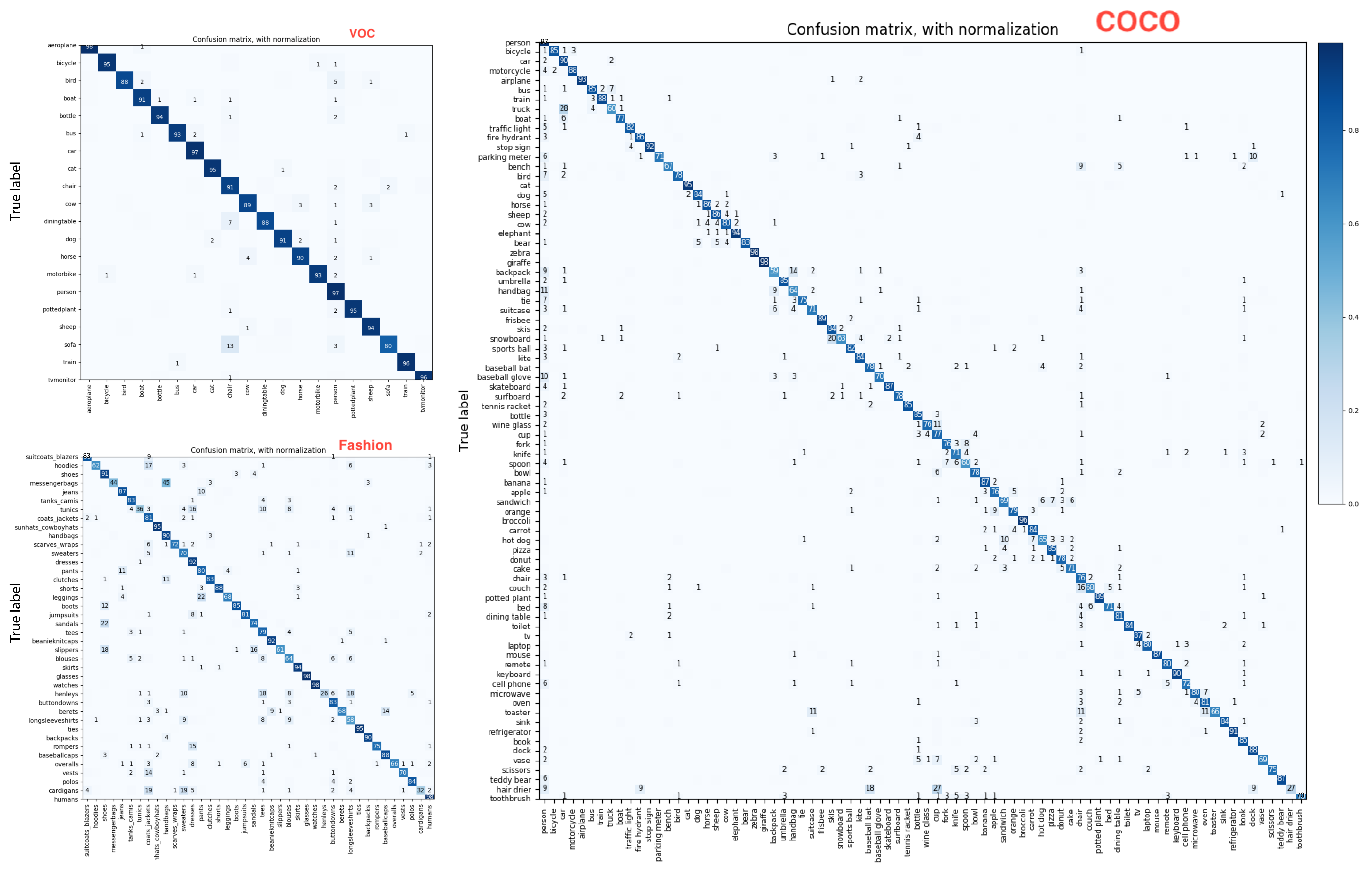}
   \caption{Confusion matrices corresponding to classifiers trained and tested on the original image size (corresponding to Table~\ref{tab:control_res}).}
\label{fig:CMs}
\end{figure*}  
\thispagestyle{empty}
\clearpage 
}

\subsection{Upper bound results}
Here, we report classification scores, upper bound APs, score of the models (mean AP over all IOUs; unless specified otherwise), and the breakdown AP over categories. 

\noindent {\bf Comparison of strategies.} Summary results of the first strategy are shown in Fig.~\ref{fig:coco_fashion_voc}. As expected UAPs over all IOUs are the same and are much better than the models. To our surprise, our second strategy did not lead to better UAP values, except for few cases including UAPs over medium and small objects on FASHION dataset and small objects on COCO (using most confident boxes), as shown in Table~\ref{tab:control_res}. Applying the original classifier, instead of training new ones on surrounding boxes, or only sampling boxes with higher IOU (\eg 0.9) did not improve the results. Also, setting the confidence of detections to 1 lowers the UAP.
We attribute the failure of the 2nd strategy to the fact that the surrounding boxes may contain additional visual content which may introduce noise in the labels. This leads to a lower classification accuracy and hence a lower AP. Therefore, in what follows we only discuss the results from the first strategy.





\begin{figure}[t]
\begin{center}
   \includegraphics[width=\linewidth]{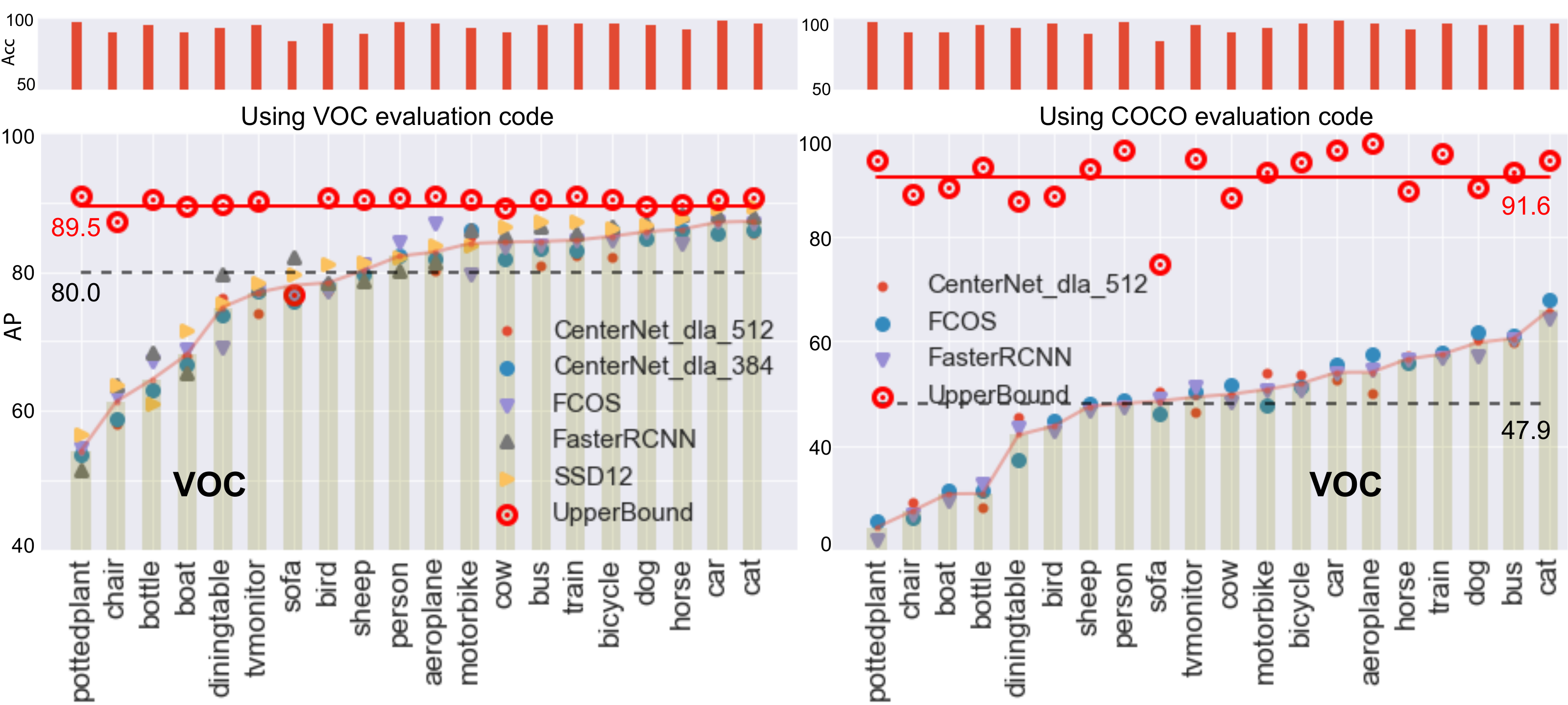}
\end{center}
   \caption{Model scores and upper bound AP over PASCAL VOC dataset using VOC (left) and COCO AP evaluation codes (right). Categories are sorted based on the average model AP. Bar charts show classification scores. Solid red and dashed black lines represent upper bound AP, and the best model AP, respectively.}
\label{fig:voc}
\end{figure}

\begin{figure}[t]
\begin{center}
   \includegraphics[width=\linewidth]{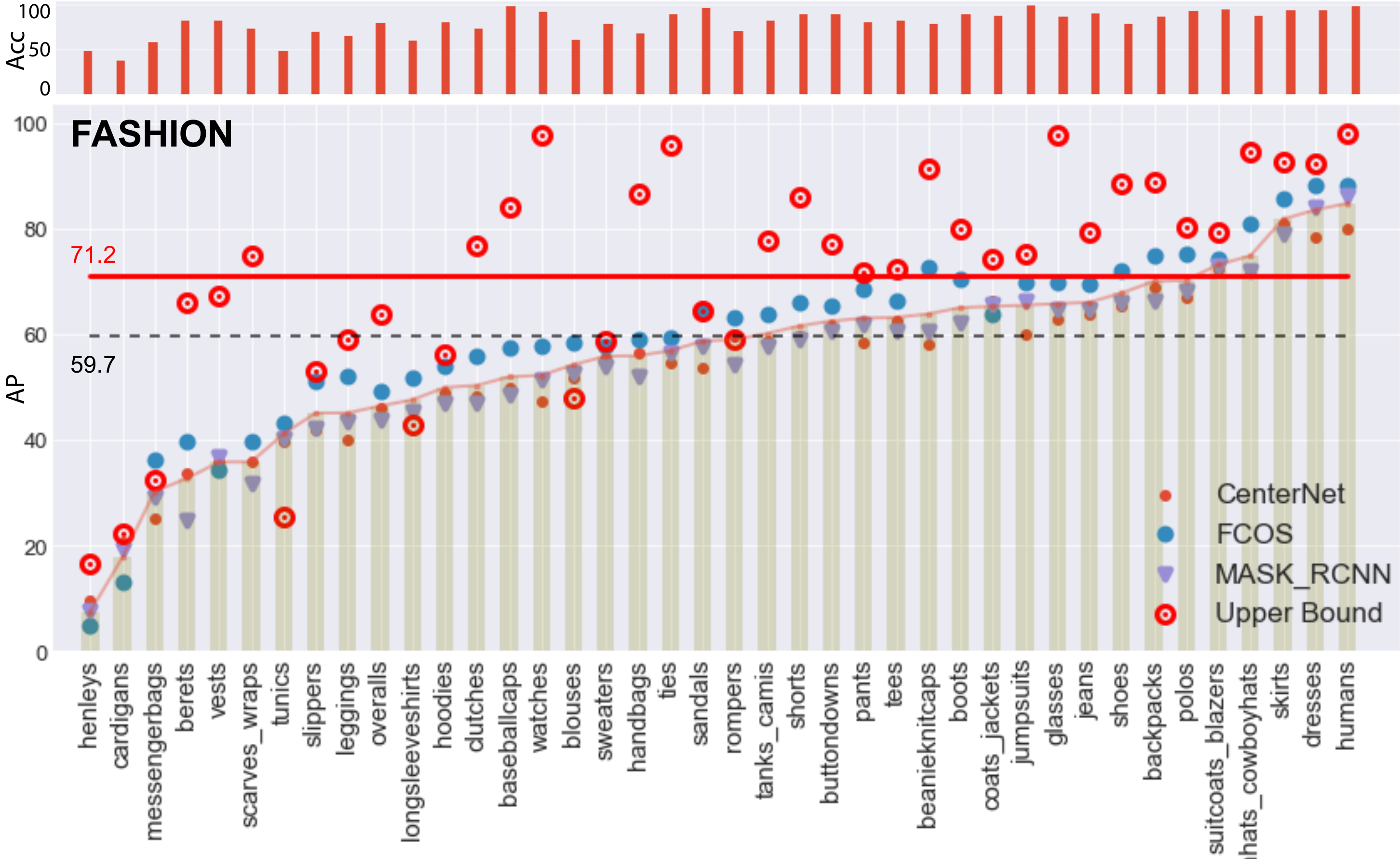}
\end{center}
   \caption{Upper bound and model APs over the FASHION dataset.}
\label{fig:fashion}
\end{figure}




\noindent {\bf PASCAL VOC}. Fig.~\ref{fig:voc} shows results using both VOC and COCO evaluation codes. The VOC evaluation code is based on IOU=0.5 and calculates the area under the PR curve slightly different than COCO. For VOC, we adopt the code from the CenterNet repository\footnote{\url{https://github.com/xingyizhou/CenterNet}}. We have trained and tested 5 models on this dataset including FasterRCNN, FCOS, SSD512, and two variants of CenterNet. The classification accuracy on VOC is very high (94.7\%). Consequently, the UAP is very high (91.6 using the COCO API). FCOS model does the best here with AP of 47.9 (right  panel in Fig.~\ref{fig:voc}; dashed lines). As it can be seen, there is a large gap between the AP of the best model and the UAP on this dataset ($\sim$45). Models are consistent in their performance across different categories.

\noindent {\bf FASHION}. Results are shown in Fig.~\ref{fig:fashion}. The best classification accuracy on this dataset is 88.8\% (Table~\ref{tab:context}). The UAP is 71.2 and the AP of the best model is 59.7 (FCOS). Interestingly, FCOS performs quite close to the upper bound at IOU=0.5 (Fig.~\ref{fig:coco_fashion_voc}). Models perform better here than over VOC. The FASHION UAP is lower than VOC UAP perhaps because classification is more challenging on the former dataset. The gap between UAP and model AP here, however, is much smaller than VOC. This could be partly due to the fact that FASHION scenes have less clutter and larger objects than the VOC scenes. While per-class UAP is above the AP of the best model over all VOC classes, UAPs of 5 FASHION categories fall below the best model AP (\emph{messenger bags, tunics, long sleeve shirts, blouses, and rompers}). Looking at the classification scores, we find that they have a low accuracy.


\begin{figure*}[htbp]
\begin{center}
   \includegraphics[width=.9\linewidth, angle=0]{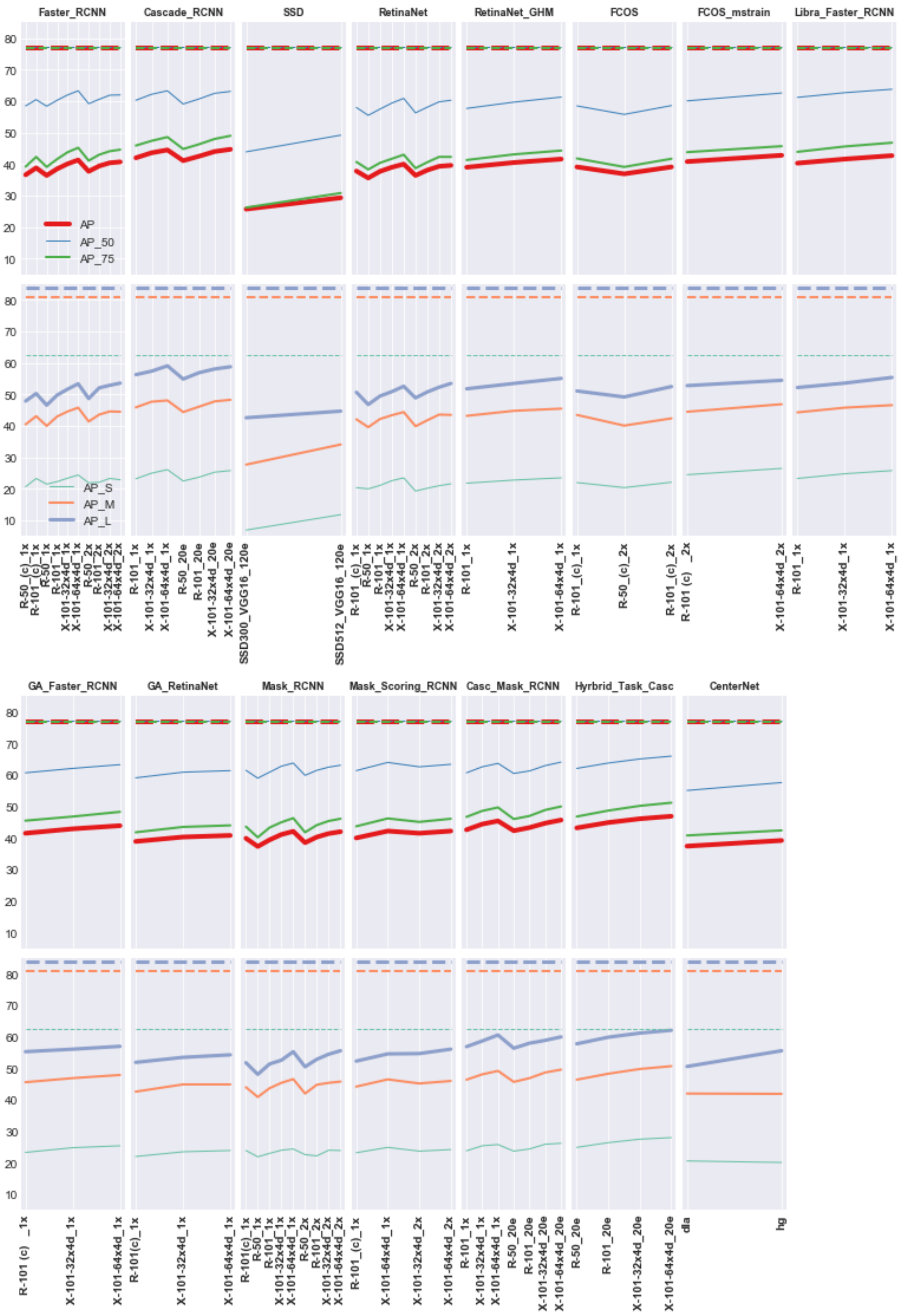}
\end{center}
   \caption{APs over COCO dataset borrowed from the \emph{MMDetection} benchmark. We add CenterNet results to \emph{MMDetection}.}
\label{fig:coco_mmdetectron}
\end{figure*}

\begin{figure*}[t]
\begin{center}
   \includegraphics[width=\linewidth]{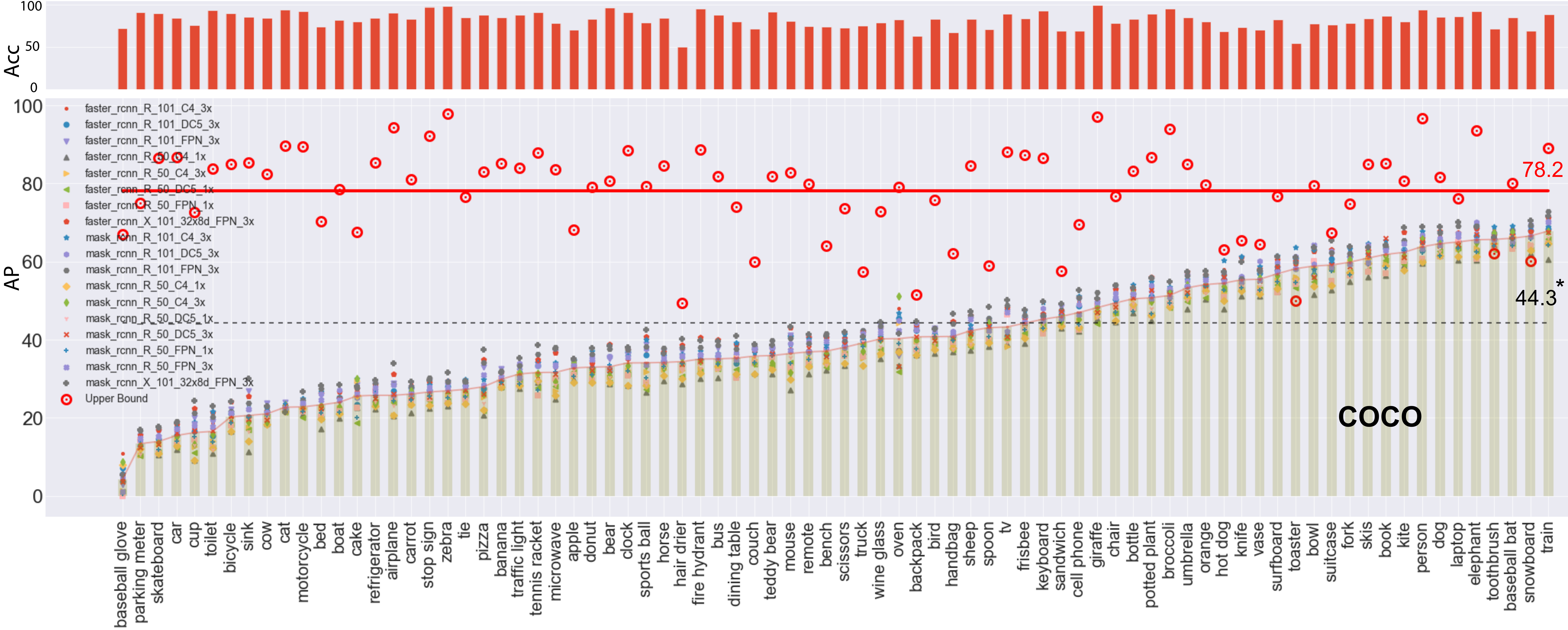}
   
\end{center}
   \caption{Detection APs over MS COCO dataset borrowed from \emph{Detectron2} benchmark. The horizontal dash line corresponds to the best model among the shown models. ``*": The best AP here is 44.3 which is smaller than the best so far on COCO (46.9). See also Fig.~\ref{fig:coco_fashion_voc}.}
\label{fig:coco_detectron}
\end{figure*}

\noindent {\bf COCO}. Existing benchmarks have provided an efficient ecosystem for developing, evaluating and comparing detection models especially on the COCO dataset. They provide trained models over a variety of settings. 
Borrowing the \emph{MMDetection} benchmark and adding the results from CenterNet to it, we end up comparing 15 models (71 in total; combination of models and backbones). Model scores are shown in Fig.~\ref{fig:coco_mmdetectron}. The best models here are Hybrid Task Cascade model~\cite{chen2019hybrid} and Cascade MaskRCNN~\cite{cai2018cascade}, with APs of 46.9 and 45.7, respectively. 
The upper bound AP on COCO is about 78.2. Recall that UAP does not depend on the IOU threshold since detected boxes are classified ground truth targets. The gap between the best model AP and UAP is above 30. The gap is much smaller for AP at IOU=0.5 which is about 10. The UAP is much lower over small objects than UAP over large objects. This also holds for models. The gap between UAP and model AP over small objects is about 35 which is much higher than the gap over medium or large objects.


Breakdown APs over object categories are shown in Fig.~\ref{fig:coco_detectron}. For this, we use the \emph{Detectron2} benchmark which reports per-category results mainly over RCNN model family. We noticed that aggregate scores on \emph{MMDetection} and \emph{Detectron2} are quite consistent.
Among 18 variants of Faster-RCNN and MASK-RCNN, the best model has the AP of 44.3 (shown by the dashed line) which is lower than the best available model on COCO (46.9; Fig.~\ref{fig:coco_fashion_voc}) and the upper bound AP. Among 80 classes, only three (\emph{snowboard, toothbrush, and toaster}) have UAPs below the best model APs.

A summary of upper bound precision and recall values over VOC, FASHION and COCO datasets is provided in Table~\ref{tab:context_res_all}.

\begin{table*}[htbp]
\begin{center}
\begin{scriptsize}
\renewcommand{\tabcolsep}{2.5pt}

\begin{tabular}{llll|ccc|ccc|ccc} 

  \multicolumn{4}{c|} {\textbf{Score}} &  \multicolumn{3}{c|} {\textbf{VOC}} &  \multicolumn{3}{c|} {\textbf{FASHION}} &  \multicolumn{3}{c} {\textbf{COCO}} \\ 

 \hline\hline
 
Avg. Prec.  & (AP) @[ IoU=0.50:0.95 & | area=   all &  | maxDets=100 ] & 0.916  & 47.3 & 47.9 & 0.712 & 0.541 & 0.597 & 0.782 & 0.364 &	0.428	\\		
Avg. Prec. &  (AP) @[ IoU=0.50    & | area=   all & | maxDets=100 ] & 0.916  & 71.3 & 71.0 & 0.712 & 0.698 & 0.711 & 0.782 & 0.584 &	0.626	\\		
Avg. Prec. &  (AP) @[ IoU=0.75    &  | area=   all & | maxDets=100 ] & 0.916  & 52.6 & 51.4 & 0.712 & 0.614 & 0.647 & 0.782 & 0.391 &	0.457	\\		
Avg. Prec. &  (AP) @[ IoU=0.50:0.95 & | area= small & | maxDets=100 ] & 0.707  & 08.6 & 11.1 & 0.457 & 0.108 & 0.182 & 0.635 & 0.215 & 0.265	\\		
Avg. Prec. &  (AP) @[ IoU=0.50:0.95 & | area=medium &  | maxDets=100 ] & 0.861  & 30.7 & 32.1 & 0.614 & 0.315 & 0.376 & 0.816 & 0.400 &	0.469	\\		
Avg. Prec. &  (AP) @[ IoU=0.50:0.95 & | area= large & | maxDets=100 ] & 0.941  & 58.1 & 58.4 & 0.721 & 0.570 & 0.627 & 0.846 & 0.466 & 0.545	\\		
Avg. Rec. &  (AR) @[ IoU=0.50:0.95 & | area=   all & | maxDets=  1 ] & 0.579  & 40.3 & 41.2 & 0.662 & 0.618 & 0.692 & 0.483 & 0.304 & 0.345	\\		
Avg. Rec. & (AR) @[ IoU=0.50:0.95 & | area=   all  & | maxDets= 10 ] & 0.908  & 53.8 & 58.5 & 0.767 & 0.712 & 0.822 & 0.797 & 0.489 &	0.552	\\		
Avg. Rec. &  (AR) @[ IoU=0.50:0.95 & | area=   all & | maxDets=100 ] & 0.930  & 54.1 & 59.5 & 0.774 & 0.714 & 0.824 & 0.812 & 0.514 & 0.582	\\		
Avg. Rec. &   (AR) @[ IoU=0.50:0.95 & | area= small &  | maxDets=100 ] & 0.736  & 11.2 & 19.5 & 0.504 & 0.194 & 0.303 & 0.663 & 0.324 & 0.388	\\		
Avg. Rec. &  (AR) @[ IoU=0.50:0.95 &  | area=medium & | maxDets=100 ] & 0.877  & 36.9 & 45.2 & 0.660 & 0.499 & 0.639 & 0.843 & 0.554 & 0.628	\\		
Avg. Rec.   &  (AR) @[ IoU=0.50:0.95 &  | area= large & | maxDets=100 ] & 0.954  & 65.7 & 70.2 & 0.782 & 0.742 & 0.850 & 0.893 &	0.645 & 0.735	\\

\end{tabular}
\end{scriptsize}
\end{center}
\caption{Precision and recall upper bounds over the three datasets (all scores).}
\label{tab:context_res_all}
\end{table*}

\noindent {\bf OpenImages}. This dataset~\cite{kuznetsova2018open} is the latest endeavor in object detection and is much more challenging than its predecessors. 
Our classifier achieves 69.0\% top-1 accuracy on the validation set of OpenImages V4 which is lower than other the three datasets. We achieve 58.9 UAP, using the TensorFlow evaluation API for computing AP\footnote{\url{https://github.com/tensorflow/models/blob/master/research/object_detection/g3doc/challenge_evaluation.md}} on this dataset, which is different than COCO AP calculation (here we discarded grouping and super-category). We are not aware of any model scores on this set of OpenImages V4.

\begin{figure}[t]
\begin{center}
   \includegraphics[width=\linewidth]{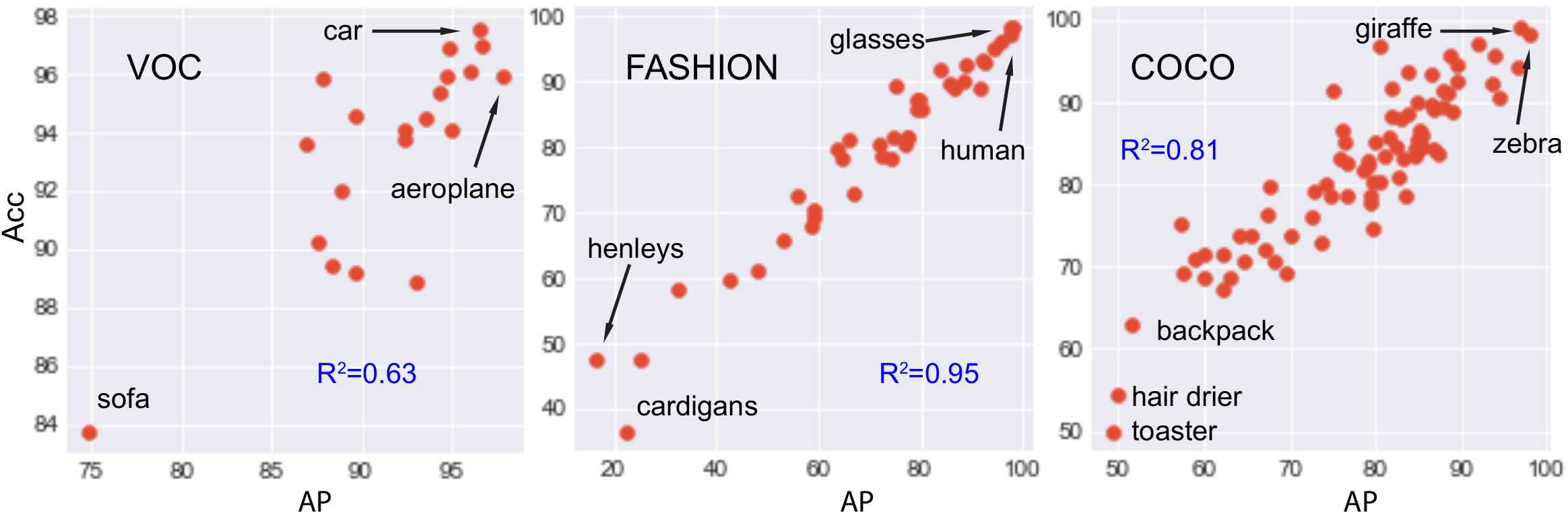}
\end{center}
   \caption{Correlation between classification accuracy and upper bound AP. The higher the Acc., the better the UAP.}
\label{fig:AP_ACC_correlation}
\end{figure}

\noindent {\bf AP vs. classification accuracy}. We found that there is a linear positive correlation ($R^2$ = 0.81 on COCO) between the UAP and the classification accuracy (Fig.~\ref{fig:AP_ACC_correlation}). The higher the classification accuracy, the higher the UAP. We did not find a correlation between the accuracy and model APs, nor between the object size and accuracy (or UAP). The dependency of UAP on accuracy, highlights the importance of recognition on object detection and constitutes the core of our analyses in the next two sections. 

\section{Error Diagnosis}
\label{sec:error}

\begin{figure}[t]
\begin{center}
   \includegraphics[width=\linewidth]{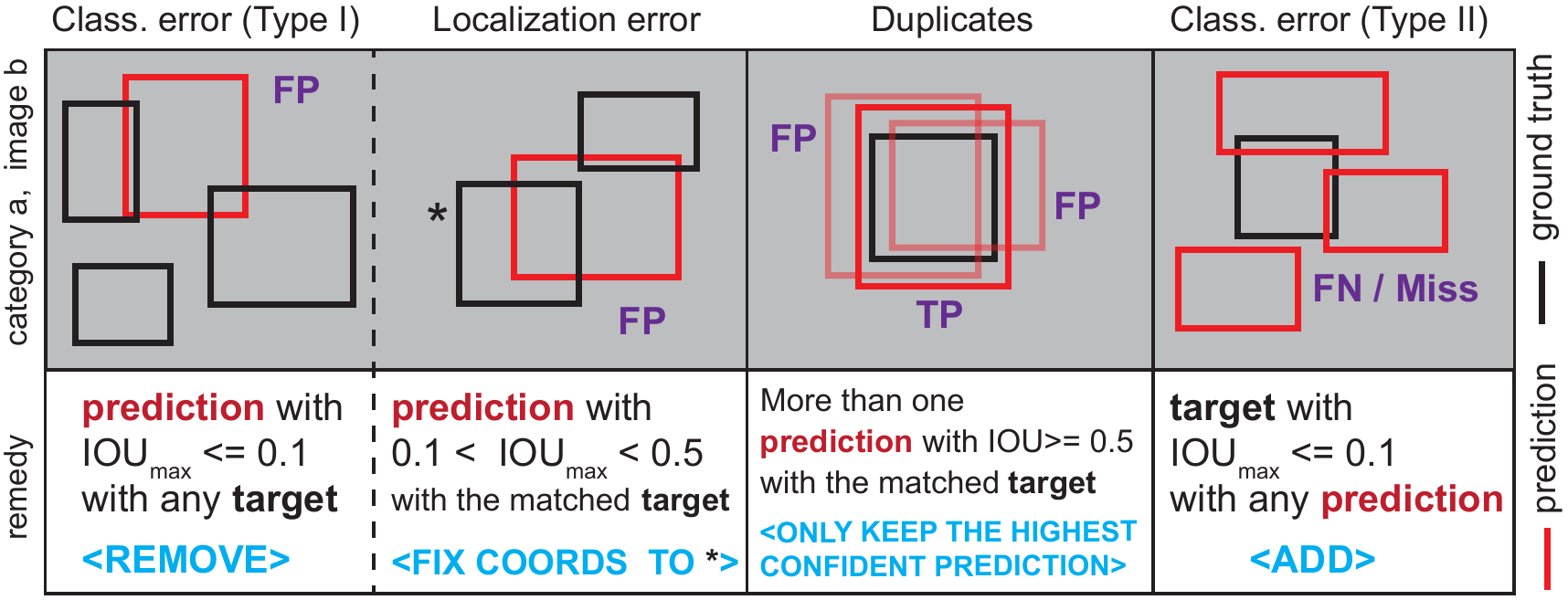}
\end{center}
   \caption{Illustration of four error types in object detection.}
\label{fig:error_types_illustration}
\end{figure}

To pinpoint the shortcomings of object detectors, we follow the analysis by Hoeim~\etal~\cite{hoiem2012diagnosing}, but revise it in two major ways. First, instead of inspecting errors across categories, here we perform a per-category error analysis (\ie binary manner). This simplifies the process and makes it easier to understand. See Fig.~\ref{fig:error_types_illustration}. We combine all types of class confusions (\eg similar classes, other classes, and background in Hoeim~\cite{hoiem2012diagnosing}) into two types of classification errors: a) \emph{confusion with the background} (Type I), and 2) \emph{misses} (Type II). Notice that this implicitly contains the above misclassification types but is much easier to investigate. In fact, recent object detectors such as FCOS~\cite{tian2019fcos} and CenterNet~\cite{zhou2019objects} also adopt this strategy to classify objects (\ie an object is of a particular class or is not). Second, Hoeim~\etal successively remove errors to reach the AP of 1. We argue that this approach convolutes different error types and does not correctly reflect the true contribution of errors (\ie understating or exaggerating error types). For example, according to the COCO analysis tool, any matches to objects with a different class label but in the same supercategory do not count as either a FP or a TP. Also, the COCO tool removes mislocalized predictions. In this case, we argue that correcting the mislocalized predictions is more effective than removing them because it can reveal other sources of weakness in a model. For example, it may lead to generating duplicates which would have been overlooked by removing the detections. 
In contrast, here we explicitly handle the errors by removing, correcting or adding detections when appropriate. Similar to Hoeim~\etal our analysis is also based on IOU=0.5.

We repeat the following procedure for each category-image pair (shown in Fig.~\ref{fig:error_types_illustration}; from left to right). First, we remove the detections with the maximum IOU$_{max}\leq0.1$ with any target (\ie classification error Type I; confusion with the background). Second, we correct the miss-localized predictions with $0.1<\text{IOU}_{max}<0.5$. In this step, coordinates of these boxes are replaced with their matching target box coordinates (which is the target with the max IOU) while their confidence scores and labels are preserved. Third, duplicates (\ie redundant detections) are removed. An unmatched detection is considered duplicate if it falls (\ie has IOU$\geq0.5$) over a target with an already assigned detection (with higher score).
Fourth, eventually, misses are treated. A miss is a target with $\text{IOU}_{max}\leq 0.1$ with any unmatched detection, and is added to the list of detections (with score of 1). Before performing this step, we set the coordinates of detections as the coordinates of their matching targets, since we now know which prediction is paired with which target (\ie one to one mapping; no duplicates). 

Results of error diagnosis are shown in Table~\ref{tab:error_type_our_results} for 3 models over 3 datasets. We start from the original detection set and progressively measure the impact of fixing each error type in the order explained above and shown in Fig.~\ref{fig:error_types_illustration}. Confusion with the background (and other classes; see above) has the highest contribution to the overall error, across all models. This indicates that models often falsely confuse background clutter or other classes as a particular object category. The second most important error type is misses. Interestingly, localization error weighs more than duplicates and has higher impact on COCO and VOC datasets than the FASHION dataset, possibly because the former two contain a larger number of small objects. Conversely, over the FASHION dataset, duplicates matter more, perhaps because class confusion is higher (\eg confusion in slippers vs. sandals; different types of hats, etc.).
Models behave almost consistently across the three datasets. 




\begin{table}[t]
\begin{center}
\begin{scriptsize}
\renewcommand{\tabcolsep}{7pt}

\begin{tabular}{l|l|c||c|c|c|c}
{\bf Dataset} & {\bf Model}      & {\bf mAP} & 

\rotatebox[origin=l]{90}{\bf - Cls. (Type I)}
 & \rotatebox[origin=l]{90}{\bf + Local.}  & \rotatebox[origin=l]{90}{\bf - Duplicates} & \rotatebox[origin=l]{90}{\bf + Misses} \\   \hline \hline
        & MaskRCNN   & 54.1              & 85.9          & 87.7         & 88.7     & 100   \\ 
FASHION   & CenterNet   & 54.0      & 88.8       & 91.7        & 96.2     & 100           \\ 
        & FCOS        & 59.7              & 90.1     & 91.9    & 95.9     & 100          \\ \Xhline{2\arrayrulewidth}
        & MaskRCNN   & 42.1              & 70.1         & 79.0           & 82.7     & 100     \\  
COCO    & CenterNet   & 39.2      & 66.1      & 78.0            & 81.7     & 100           \\ 
        & FCOS        & 42.8      & 69.6      & 80.8             & 85.4     & 100           \\ \Xhline{2\arrayrulewidth}
        & MaskRCNN & 47.3     & 73.7        & 78.8     & 79.7     & 100           \\ 
VOC  & CenterNet   & 47.8              & 79.0      & 88.5           & 92.6     & 100           \\ 
        & FCOS        & 47.9      & 76.3    & 85.0               & 90.3     & 100           \\          
\end{tabular}
\end{scriptsize}
\end{center}
\caption{Quantifying the contribution of errors in object detection. ``Local." and ``Dup." stand for localization error and duplicate removal, respectively. mAP is the model AP over all IOUs. }
\label{tab:error_type_our_results}
\end{table}

\begin{figure}[t]
\begin{center}
   \includegraphics[width=\linewidth]{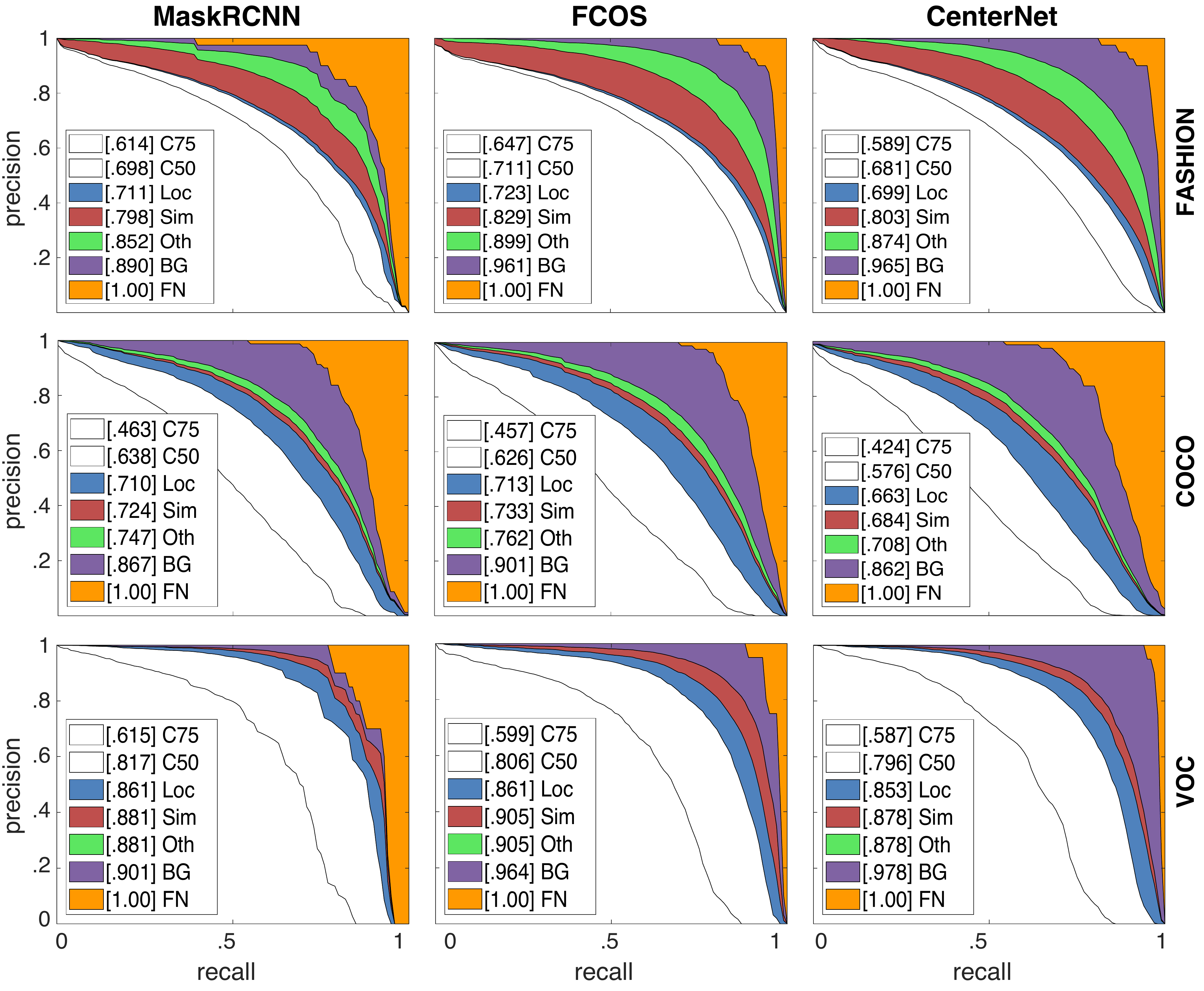}
\end{center}
   \caption{Quantifying the contribution of errors in object detection using the MS COCO analysis code.}
\label{fig:error_type_all_results}
\end{figure}  

\begin{figure}[t]
\begin{center}
   \includegraphics[width=\linewidth]{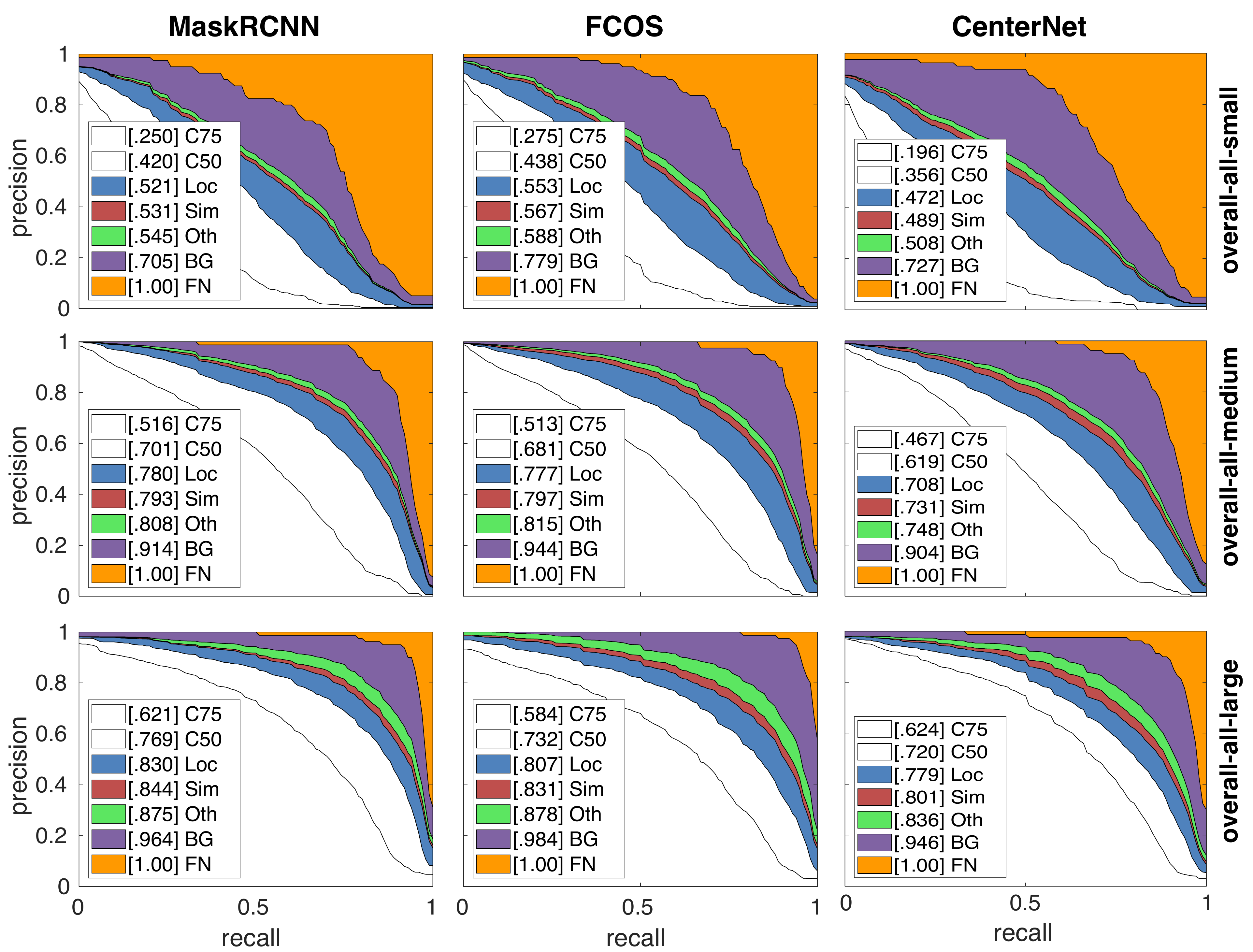}
\end{center}
   \caption{Error analysis of models based on the object size over the COCO dataset.}
\label{fig:error_type_coco_results}
\end{figure}

We also cross checked our results with results obtained using the COCO analysis tool (implementing Hoeim~\etal). Notice that numbers from COCO analysis tool are not directly comparable to ours since our strategy is different and, unlike us, it does not explicitly address duplicate errors. Nevertheless, based on APs and PR curves in Fig.~\ref{fig:error_type_all_results}, we arrive at similar conclusions to ours. Here, again we observe that classification error Type I (Sim, Oth, and BG in Fig.~\ref{fig:error_type_all_results}) accounts for the largest fraction of errors, followed by misses (FN) and localization (Loc) errors. Fig.~\ref{fig:error_type_coco_results} shows the breakdown of error analysis over small, medium, and large objects using the COCO analysis tool. As it can be seen, all three models miss a much larger number of small objects compared to medium or large ones. As expected, models obtain a much higher mAP over large objects than small ones. A lot or background regions, however, are still classified as large objects.

\section{Invariance Analysis}
\label{sec:invariance}
\begin{figure}[t]
\begin{center}

   \includegraphics[width=\linewidth]{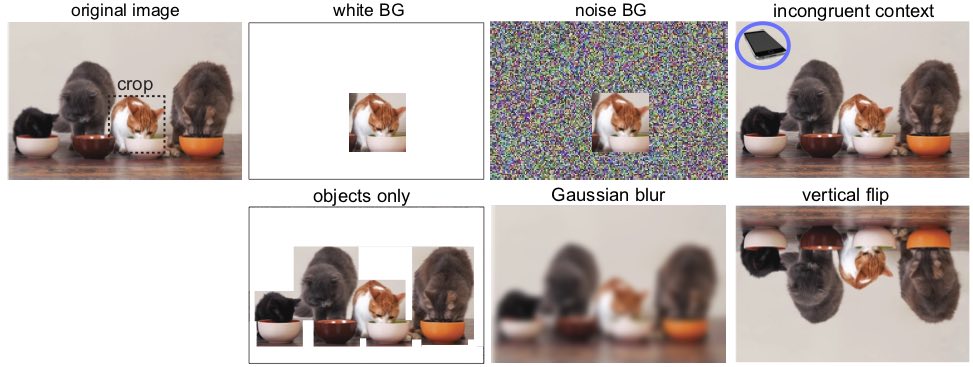}
\end{center}
   \caption{Illustration of the experiments regarding invariance analysis of object detection approaches.}
\label{fig:invariance1}
\end{figure}

\begin{figure*}[htbp]
\begin{center}
   \includegraphics[width=1\linewidth, angle=0]{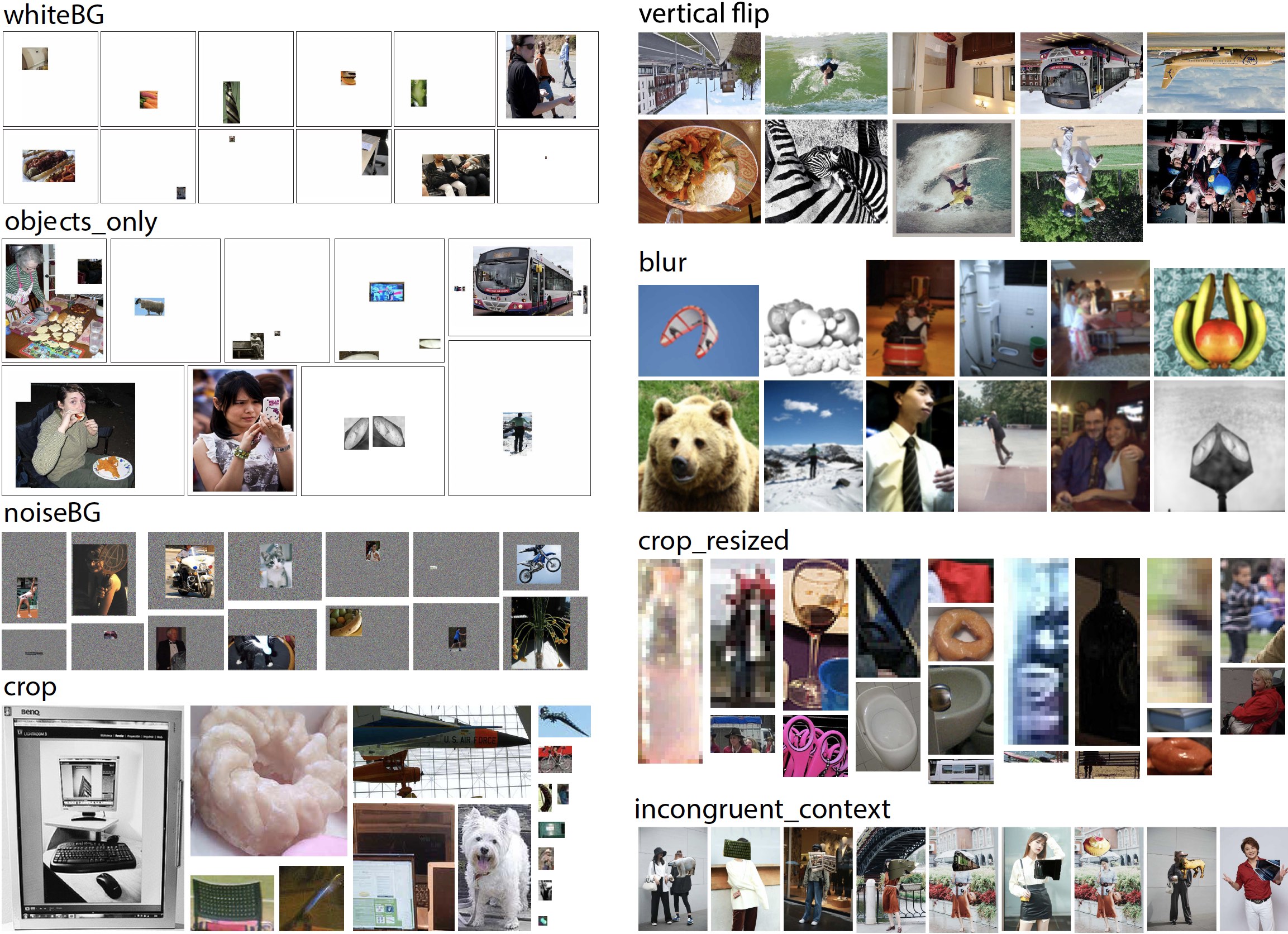}
\end{center}
   \caption{Sample images used in the invariance analysis experiments. Left column) single objects in white background, objects\_only (\ie removing background), single objects in white noise background, cropped objected (no resizing). Right column) vertically flipped images, Gaussian blurred images, cropped and resized objects (width=300), and objects in incongruent contexts.}
\label{fig:collage}
\end{figure*}

Complementary to our error diagnosis, here we conduct a series of experiments to reduce the impact of localization or recognition in detection pipelines (one at a time). Our principal emphasize is on the recognition component. These experiments are performed over the \emph{COCOval2017} set and are illustrated in Fig.~\ref{fig:invariance1} and Fig.~\ref{fig:collage}. Trained models, over the \emph{COCOtrainval0712} set, are employed.

\begin{figure}[htbp]
\begin{center}
   \includegraphics[width=.7\linewidth]{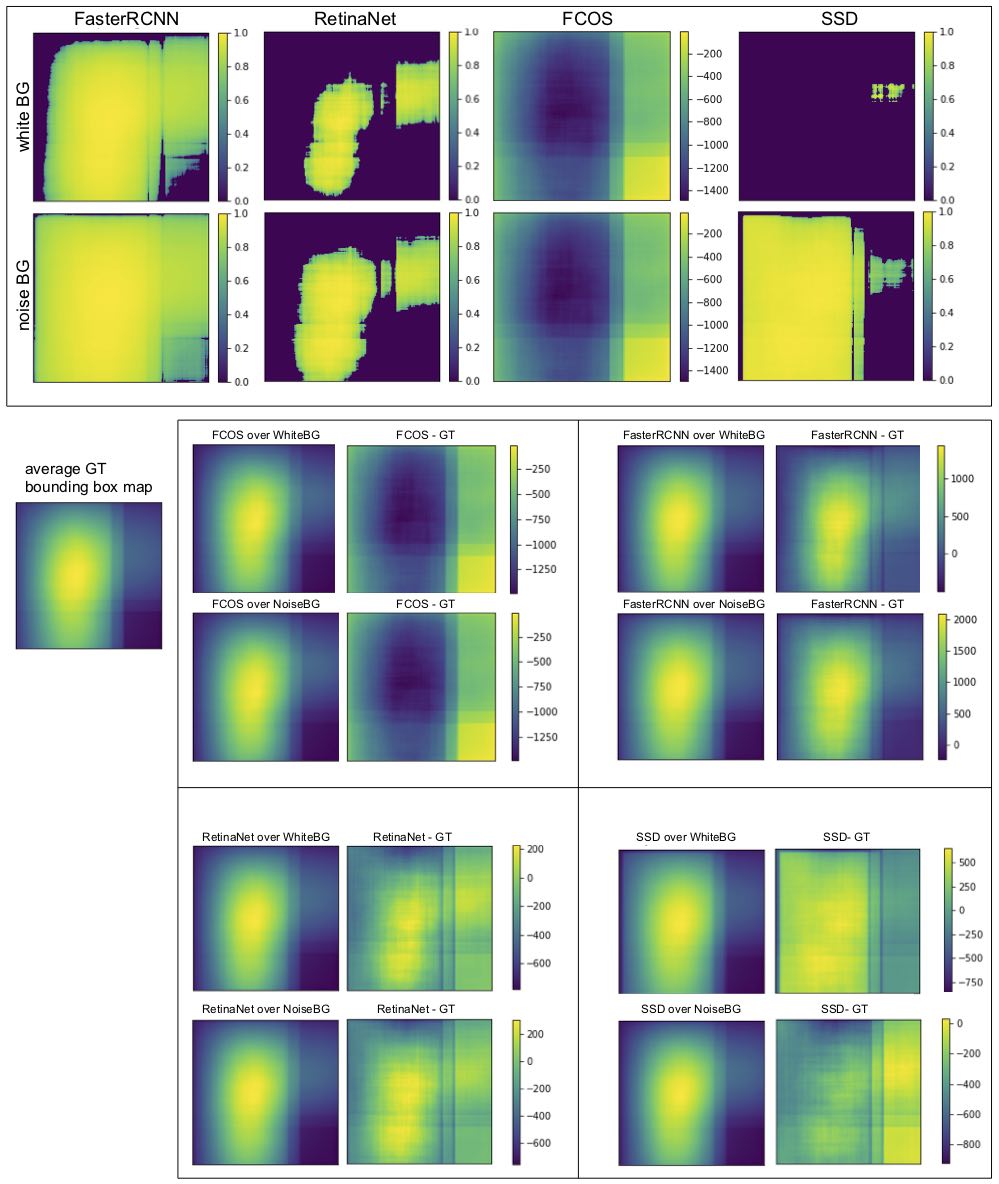}
\end{center}
   \caption{Top) Results of invariance analysis of models over whiteBG and noiseBG images. It shows the difference in the distribution of predicted boxes (spaces delimited by boxes are superimposed) by a model and the distribution of ground-truth predicted boxes (in log scale). Bottom) Same as above but in linear scale. 
   The left (alone) panel shows the distribution of target object boxes in whiteBG (same as noiseBG) images over the \emph{COCOval2017} dataset. Second column in each sub panel (in the right panel) shows the difference in the distributions of predicted vs. ground-truth boxes.}
\label{fig:invariance2}
\end{figure}

{\bf Analysis of context.}
In the first experiment, we generated stimuli in which a single object was placed in a white background or in a white noise background (one object per image, hence number of images equal to the number of objects). Contrary to our expectation, 
we found that models either underestimate or overestimate the distribution of target bounding boxes. Table~\ref{tab:num_detected_box} shows the number of generated bounding boxes by models. All models overestimate the number of ground-truth bounding boxes which is 36,781. Interestingly, FasterRCNN generates a significantly lower number of boxes compared to other models. Fig.~\ref{fig:invariance2} shows the difference in distribution of predicted boxes and distribution of ground-truth boxes. Interestingly, models search all over the place. FasterRCNN and RetinaNet oversample boxes around targets, while FCOS generates a fair amount. This hints towards the shortcomings in objectness prediction in models.
Quantitative results, presented in Table~\ref{tab:invariance_results1}, show that models perform poorly on these images (about the same in both conditions but lower than the original images). They are hindered much more on small objects than medium or large ones, which shows how critical context is for recognition and detection of small objects. Interestingly, in white/noise BG and object-only cases, the AP-large increases but the AP-small decreases (compared to orig. images). FCOS, ranking higher on original images, does better here as well. FCOS, ranking higher on original images, does better here as well.

\begin{table}
\begin{center}
\renewcommand{\tabcolsep}{5pt}

\begin{tabular}{l|ccccc|ccccc|ccc}

        & {\bf RetinaNet} & {\bf FasterRCNN} & {\bf SSD} & {\bf FCOS} \\ \hline \hline
whiteBG & 1,029,369  & 209,094  & 2,236,791  & 1,302,592 \\ 
noiseBG & 1,540,733 & 214,769 & 3,570,051 & 2,320,184 \\


\end{tabular}
\end{center}
\caption{Number of generated boxes on the whiteBG and noiseBG cases by different models. Notice that the number of ground-truth bounding boxes are 36,781 (validation set of COCO; here one object per image).}
\label{tab:num_detected_box}
\end{table}

\begin{table}
\begin{center}
\begin{scriptsize}
\renewcommand{\tabcolsep}{7pt}

\begin{tabular}{l|ccc|ccc|ccc}
\multirow{2}{*}{\textbf {Model}} & \multicolumn{3}{c|} {\textbf{white BG}} & 
\multicolumn{3}{c|} {\textbf{noise BG}}& 
\multicolumn{3}{c} {\textbf{objects\_only}} \\
 \cline{2-10} 
& $AP$ & $AP^{.5}$ & $AP^{.75}$ & $AP$ & $AP^{.5}$ & $AP^{.75}$ & $AP$ & $AP^{.5}$ & $AP^{.75}$ \\
\hline\hline
FasterRCNN & 31.1 & 42 & 36.1 & 31.8 & 39.8 & 36.8 & 35.9 & 55.8 & 39.5 \\
RetinaNet & 33.1 & 41.0 & 37.3 & 32.7 & 39.1 & 36.6 & 39.8 & 58.4 & 43.4 \\
FCOS & 34.5 & 42 & 37.1 & 34.2 & 39.8 & 37.4 &   43.6 & 60.6 & 46.9 \\
SSD512 & 27.4 & 36.7 & 32.3 & 26.0 & 33.4 & 34 &  30.5 & 48.6 & 32.9 \\
\Xhline{2\arrayrulewidth}
\textbf {Model} & $AP_s$ & $AP_m$ & $AP_l$ & $AP_s$ & $AP_m$ & $AP_l$ & $AP_s$ & $AP_m$ & $AP_l$ \\
\hline
FasterRCNN & 7.5 & 35.9 & 49.9 & 7.0 & 36.6 & 52.1 & 17.5 & 40.6 & 48.6 \\
RetinaNet & 8.3 & 37.5 & 53.2 & 6.4 & 38.3 & 54.2 & 18.9 & 44.5 & 56.4 \\
FCOS & 8.5 & 39.8 & 55.2 & 9.4 & 39.5 & 54.8 & 22.1 & 48.8 & 58.7  \\
SSD512 & 7.0 & 31.4 & 45.1 & 4.6 & 29.3 & 45.2 &  9.8 & 35.7 & 48.4 \\

\end{tabular}
\end{scriptsize}
\end{center}
\caption{Results of invariance analysis over \emph{COCOval2017}.}
\label{tab:invariance_results1}
\end{table}


In the second experiment, object-only case, we removed the image background and preserved all the objects (hence the same number of images as in \emph{COCOval2017}). To our surprise, FCOS and SSD performed better on these images than the original ones (Column 1 vs. 10 in Table~\ref{tab:invariance_results1}). Compared to the original images, they did better on large objects and lower on small objects in the object-only case.

In the third experiment, we paste objects in incongruent backgrounds (\eg a boat in the street), similar to~\cite{rosenfeld2018elephant} but over a larger dataset and including more models. Also, unlike Rosenfeld~\etal, we report the AP. We paste 9 objects including \emph{bear, keyboard, refrigerator, surfboard, train, tv, cake, horse, and oven} on 100 images taken from the FASHION dataset; 900 images in total. Fig.~\ref{fig:collage} shows some examples. Results are given in Table.~\ref{tab:context_incongruent_results}. Interestingly, models performed well on this dataset. They failed drastically on \emph{surfboard} and \emph{oven} which seem to be a little hard for humans. \emph{Cake, bear, and horse} were the easiest ones. FCOS did the best among models. Overall, we did not a dramatic failure of models in detecting out of context objects, at least on the set of objects we tried. Nonetheless, in some other scenarios (\eg smaller objects) models may fail to detect objects out of their common contexts. This highlights and aligns with the current view that deep learning models fit themselves to the statistics of the datasets. We believe that retraining the object detectors on these examples can alleviate the problem to some extent. Similar attempts have been made in the past to strengthen object detectors by training them on degraded images (\eg as in Michaelis~\etal~\cite{michaelis2019benchmarking}) or making recognition models robust to adversarial examples through adversarial training (\eg as in Goodfellow~\etal~\cite{goodfellow2014explaining}).

\begin{table}
\begin{center}
\begin{small}
\renewcommand{\tabcolsep}{3pt}

\begin{tabular}{l|ccccccccc|c}
Model & \rotatebox[origin=l]{90}{train} & \rotatebox[origin=l]{90}{horse} & \rotatebox[origin=l]{90}{bear} & \rotatebox[origin=l]{90}{surfboard} & \rotatebox[origin=l]{90}{cake} & \rotatebox[origin=l]{90}{tv} & \rotatebox[origin=l]{90}{keyboard} & \rotatebox[origin=l]{90}{oven} & \rotatebox[origin=l]{90}{fridge} & Avg. \\

\hline
\hline

Fa.RCNN & 64.0 & 58.4 & 84.7 & 2.4 & 77.9 & 74.3 & 54.7 & 15.5 & 20.3 & 50.2 \\
RetinaNet  & 54.2& 89.2& 90.6& 2& 85.7& 86.6& 10.1& 24.8& 69.3 & 57.0\\
FCOS       & 73.4& 91.5& 94.0 & 17.1& 87.6& 92.1& 9.8& 44.2& 76.2 & {\bf 65.1}\\
SSD512 .   & 84.3& 58.9& 78.5& 3.8& 76.9& 69.8& 42.6& 8.4& 47.6 & 52.3\\ 
\Xhline{2\arrayrulewidth}
Avg. & 69.0& 74.5  & {\bf 87.0} & 6.3 & 82.0 & 80.7 & 29.3 & 23.2 & 53.4 & 56.2

\end{tabular}
\end{small}
\end{center}
\caption{Model APs (IOU=.5) over objects in incongruent contexts. FCOS does better than other models.}
\label{tab:context_incongruent_results}
\end{table}


{\bf Robustness to image transformations.}
In the fourth experiment, we evaluated models on objects that were a) cropped right out of the image, or b) cropped and resized such that their smallest dimension became 300 pixels (while preserving the aspect ratio). Models performed terribly in both cases as shown in Table~\ref{tab:invariance_results2}, with RetinaNet doing better. Poor performance here demonstrates how sensitive models are to object scale and that they lack robustness to object appearance. Visually inspecting the images, we found it very difficult to recognize the cropped objects, especially the small ones.

Fifth and sixth experiments regard testing models on 
Gaussian blur (with a 11 $\times$ 11 kernel) and vertical flip, respectively. Results in Table~\ref{tab:invariance_results2} 
show that both types of transformations dramatically hinder performance with higher impact for vertical flip. We do not have a baseline for human performance on these cases, but a quick browsing shows that it is still possible to detect objects, albeit with more effort. RetinaNet and FCOS outperform other models here. 

\noindent {\bf Analysis of errors.} Here we measure the impact of each error type in three detection tasks including object-only, Gaussian blur and vertical flip. See Table~\ref{tab:error_type_invariance_results} for results. Error types in order of importance include: \emph{misses, localization, misclassification (Type I), and duplicates}, over three tasks. Models miss more objects in vertical flip and Gaussian blur cases compared to the objects-only case. There is less confusion with BG in objects-only case than original images (classification Type I) since there is no background clutter.

Finally, Table~\ref{tab:invariance_all} and Table~\ref{tab:invariance_category} summarize all results of the invariance analysis experiments, including precision, recall, and breakdown over categories.

\begin{figure}[htbp]
\begin{center}
   \includegraphics[width=.85\linewidth, angle=0]{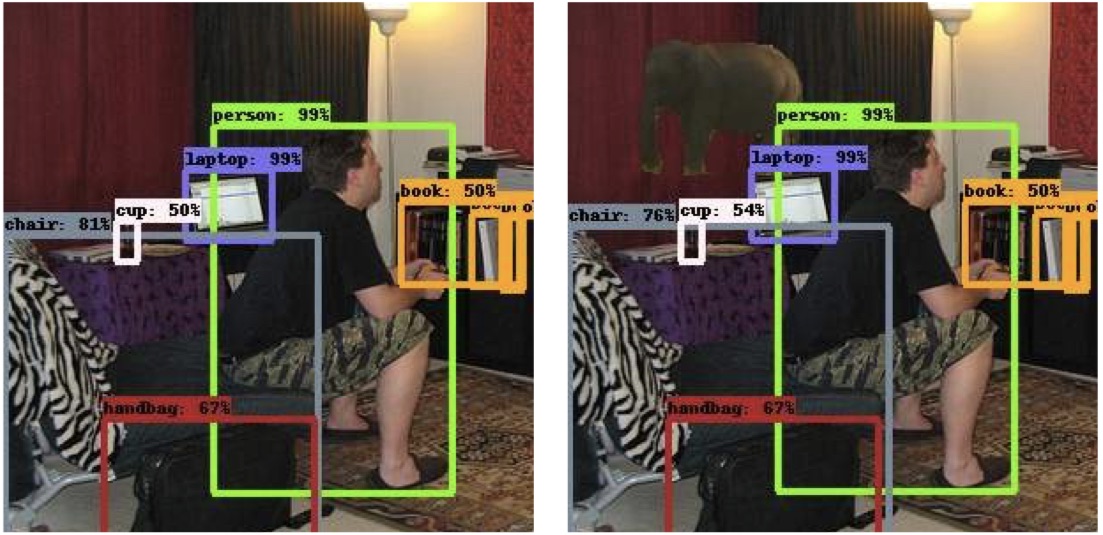}
\end{center}
   \caption{A state-of-the-art object detector (Faster-RCNN; trained on COCO dataset) is able to detects multiple objects in a living-room (a), but it fails to detect a transplanted object (elephant) out of its context. Rosenfeld \etal~\cite{rosenfeld2018elephant} showed that a transplanted object a) may occasionaly become undetected or be detected with sharp changes in confidence, b) may be classified as another object, and/or c) cause other objects to switch identity, bounding box, or disappear (image reproduced from~\cite{rosenfeld2018elephant}).}
\label{fig:elephant}
\end{figure}

\begin{table*}
\begin{center}
\begin{scriptsize}
\renewcommand{\tabcolsep}{8pt}

\begin{tabular}{l|ccc|ccc|ccc|cc}
\multirow{2}{*}{\textbf {Model}} & \multicolumn{3}{c|} {\textbf{crop}} & 
\multicolumn{3}{c|} {\textbf{Gaussian blur}}& 
\multicolumn{3}{c|} {\textbf{vertical flip}} & \multicolumn{2}{c} {\textbf{orig img.}}\\
 \cline{2-12} 
& $AP$ & $AP^{.5}$ & $AP^{.75}$ & $AP$ & $AP^{.5}$ & $AP^{.75}$ & $AP$ & $AP^{.5}$ & $AP^{.75}$ & $AP$ & $AP^{.5}$ \\
\hline\hline
FasterRCNN &  8.4 & 15.0 & 8.2 &  17.1 & 29.6 & 17.4 & 15.5 & 27.3 & 15.7 & 36.4 & 58.4\\
RetinaNet & 16.9 & 22.7 & 18.8 &  21.5 & 34.7 & 22.5 & 18.7 & 30.7 & 19.3 & 40.0 & 60.9\\
FCOS & 14.3 & 18.5 & 15.3 &  21.0 & 33.7 & 21.6 & 19.1 & 30.2 & 19.6 & 42.8 & 62.6 \\
SSD512 & 13.4 & 18.9 & 14.9 & 15.1 & 26.6 & 15.2 & 12.1 & 22.2 & 11.9 & 29.3 & 49.2 \\
\Xhline{2\arrayrulewidth}
\textbf {Model} & $AP_s$ & $AP_m$ & $AP_l$ & $AP_s$ & $AP_m$ & $AP_l$ & $AP_s$ & $AP_m$ & $AP_l$ & $AP_s$ & $AP_l$\\
\hline
FasterRCNN & 0 & 1.3 & 18.7 &     3.8 & 18.3 & 31.5 & 6.2 & 16.6 & 24.7 & 21.5 & 46.6\\
RetinaNet &  1 & 5.2 & 34.1 &    5.1 & 22.8 & 39.0 &  7.5 & 20.5 & 29.5 & 23.5 & 52.6 \\
FCOS & 1 & 4.5 & 32.2 &  5.3 & 22.5 & 37.4 & 8.0 & 20.8 & 30.0 & 26.5 & 54.5 \\
SSD512 & 1 & 2.9 & 25.7 &  2.0 & 15.2 & 30.9 & 4.0 & 12.6 & 22.5 & 11.8 & 44.7 \\

\end{tabular}
\end{scriptsize}
\end{center}
\caption{Additional results of invariance analysis over \emph{COCOval2017} dataset. }
\label{tab:invariance_results2}
\end{table*}

\begin{table}
\begin{center}
\begin{small}
\renewcommand{\tabcolsep}{8pt}

\begin{tabular}{l|l|c||c|c|c|c}
{\bf Dataset} & {\bf Model}      & {\bf mAP} &\rotatebox[origin=l]{90}{\bf - Cls. (Type I)}
 & \rotatebox[origin=l]{90}{\bf + Local.}  & \rotatebox[origin=l]{90}{\bf - Duplicates} & \rotatebox[origin=l]{90}{\bf + Misses} \\   \hline \hline
objects               & Fa.RCNN & 55.8              & 61.5                       & 69.3                               & 75.2     & 100           \\ 
only  & RetinaNet & 58.4              & 64.6                       & 72.6                                & 79.9     & 100           \\ 
               & FCOS       & 60.6              & 67.8                       & 77.0                                & 82.3     & 100           \\ \Xhline{2\arrayrulewidth}
Gaussian               & Fa.RCNN  & 29.6              & 37.2                       & 47.4                                & 55.2     & 100           \\ 
 blur          & RetinaNet & 34.7              & 42.3                       & 53.5                                & 64.3     & 100           \\ 
               & FCOS       & 33.7              & 43.1                       & 56.8                                & 65.3     & 100           \\ \Xhline{2\arrayrulewidth}
vertical               & Fa.RCNN  & 27.3              & 37.0                       & 49.6                                & 57.3     & 100           \\ 
 flip & RetinaNet & 30.7              & 41.1                       & 54.1                                & 64.6     & 100           \\ 
               & FCOS       & 30.2              & 41.3                       & 57.1                                & 65.6     & 100          \\ 
            
\end{tabular}
\end{small}
\end{center}
\caption{Error analysis of models over transformed images. }
\vspace{-10pt}
\label{tab:error_type_invariance_results}
\end{table}

\begin{table*}[htbp]
\begin{center}
\begin{scriptsize}
\renewcommand{\tabcolsep}{3pt}
\renewcommand\arraystretch{1.1}

\begin{tabular}{c|llll|lllllll|l|} 

 \textbf{Model} &  \multicolumn{4}{c|} {\textbf{Score}} & \rotatebox[origin=l]{90}{whiteBG}  & \rotatebox[origin=l]{90}{noiseBG} & \rotatebox[origin=l]{90}{crop\_resized} & \rotatebox[origin=l]{90}{crop} & \rotatebox[origin=l]{90}{objects\_only} & \rotatebox[origin=l]{90}{blur} & \rotatebox[origin=l]{90}{vertical\_flip} & \rotatebox[origin=l]{90}{incongruent\_context} \\
 
 \hline\hline

\multirow{12}{*} {\textbf{\rotatebox[origin=c]{90}{FasterRCNN}}} &  Avg. Prec. (AP) 	&	@[ IoU=50:95 	&	| area=  all 	&	| maxDets=100 ]	&	33.1	&	32.7	&	14.3	&	0.084	&	0.398	&	0.215	&	0.187	&	0.569	\\
& Avg. Prec. (AP) 	&	@[ IoU=50	&	| area=  all 	&	| maxDets=100 ]	&	41	&	39.1	&	19.4	&	0.15	&	0.584	&	0.347	&	0.307	&	0.659	\\
& Avg. Prec. (AP) 	&	@[ IoU=75	&	| area=  all 	&	| maxDets=100 ]	&	37.3	&	36.6	&	15.9	&	0.082	&	0.434	&	0.225	&	0.193	&	0.636	\\
& Avg. Prec. (AP) 	&	@[ IoU=50:95	&	| area= small 	&	| maxDets=100 ]	&	8.3	&	6.4	&	-1	&	0	&	0.189	&	0.051	&	0.075	&	-1	\\
& Avg. Prec. (AP) 	&	@[ IoU=50:95	&	| area=medium 	&	| maxDets=100 ] 	&	37.5	&	38.3	&	0.001	&	0.013	&	0.445	&	0.228	&	0.205	&	-1	\\
& Avg. Prec. (AP) 	&	@[ IoU=50:95	&	| area= large  	&	| maxDets=100 ]	&	53.2	&	54.2	&	16.1	&	0.187	&	0.564	&	0.39	&	0.295	&	0.571	\\
& Avg. Rec.  (AR)	&	@[ IoU=50:95	&	| area=  all  	&	| maxDets= 1 ]	&	55	&	54.3	&	31.7	&	0.214	&	0.335	&	0.223	&	0.216	&	0.67	\\
& Avg. Rec.  (AR)	&	@[ IoU=50:95	&	| area=  all 	&	| maxDets= 10 ]	&	57.1	&	56.9	&	35.5	&	0.254	&	0.52	&	0.342	&	0.349	&	0.681	\\
& Avg. Rec.  (AR)	&	@[ IoU=50:95	&	| area=  all  	&	| maxDets=100 ]	&	57.2	&	56.9	&	35.7	&	0.255	&	0.549	&	0.358	&	0.369	&	0.681	\\
& Avg. Rec.  (AR)	&	@[ IoU=50:95	&	| area= small  	&	| maxDets=100 ]	&	25.1	&	22	&	-1	&	0.045	&	0.309	&	0.089	&	0.152	&	-1	\\
& Avg. Rec.  (AR)	&	@[ IoU=50:96	&	| area=medium 	&	| maxDets=100 ]	&	68.4	&	70.6	&	6.8	&	0.164	&	0.608	&	0.385	&	0.384	&	-1	\\
& Avg. Rec.  (AR)	&	@[ IoU=50:97	&	| area= large 	&	| maxDets=100 ]	&	81.4	&	83.2	&	35.8	&	0.435	&	0.73	&	0.625	&	0.577	&	0.681	\\
\hline																							
\multirow{12}{*} {\textbf{\rotatebox[origin=c]{90}{RetinaNet}}} & Avg. Prec. (AP) 	&	@[ IoU=50:95 	&	| area=  all 	&	| maxDets=100 ]	&	31.1	&	31.8	&	11.2	&	0.169	&	0.359	&	0.21	&	0.155	&	0.502	\\
& Avg. Prec. (AP) 	&	@[ IoU=50	&	| area=  all 	&	| maxDets=100 ]	&	40.2	&	39.8	&	16.9	&	0.227	&	0.558	&	0.337	&	0.273	&	0.715	\\
& Avg. Prec. (AP) 	&	@[ IoU=75	&	| area=  all 	&	| maxDets=100 ]	&	36.1	&	36.8	&	12.1	&	0.188	&	0.395	&	0.216	&	0.157	&	0.579	\\
& Avg. Prec. (AP) 	&	@[ IoU=50:95	&	| area= small 	&	| maxDets=100 ]	&	7.5	&	7	&	-1	&	0.001	&	0.175	&	0.053	&	0.062	&	-1	\\
& Avg. Prec. (AP) 	&	@[ IoU=50:95	&	| area=medium 	&	| maxDets=100 ] 	&	35.9	&	36.6	&	0.005	&	0.052	&	0.406	&	0.225	&	0.166	&	-1	\\
& Avg. Prec. (AP) 	&	@[ IoU=50:95	&	| area= large  	&	| maxDets=100 ]	&	49.9	&	52.1	&	13.2	&	0.341	&	0.486	&	0.374	&	0.247	&	0.511	\\
& Avg. Rec.  (AR)	&	@[ IoU=50:95	&	| area=  all  	&	| maxDets= 1 ]	&	47	&	48.8	&	21.8	&	0.396	&	0.302	&	0.222	&	0.187	&	0.575	\\
& Avg. Rec.  (AR)	&	@[ IoU=50:95	&	| area=  all 	&	| maxDets= 10 ]	&	48.5	&	50.1	&	24.5	&	0.452	&	0.474	&	0.344	&	0.294	&	0.596	\\
& Avg. Rec.  (AR)	&	@[ IoU=50:95	&	| area=  all  	&	| maxDets=100 ]	&	48.5	&	50.1	&	24.6	&	0.454	&	0.495	&	0.357	&	0.307	&	0.596	\\
& Avg. Rec.  (AR)	&	@[ IoU=50:95	&	| area= small  	&	| maxDets=100 ]	&	16.1	&	15.8	&	-1	&	0.099	&	0.266	&	0.094	&	0.119	&	-1	\\
& Avg. Rec.  (AR)	&	@[ IoU=50:96	&	| area=medium 	&	| maxDets=100 ]	&	58.7	&	61.3	&	10	&	0.459	&	0.554	&	0.385	&	0.322	&	-1	\\
& Avg. Rec.  (AR)	&	@[ IoU=50:97	&	| area= large 	&	| maxDets=100 ]	&	73.9	&	77.4	&	24.6	&	0.688	&	0.657	&	0.615	&	0.492	&	0.596	\\
\hline		

\multirow{12}{*} {\textbf{\rotatebox[origin=c]{90}{FCOS}}} & Avg. Prec. (AP) 	&	@[ IoU=50:95 	&	| area=  all 	&	| maxDets=100 ]	&	34.5	&	34.2	&	12.1	&	0.143	&	0.436	&	0.171	&	0.191	&	0.651	\\
& Avg. Prec. (AP) 	&	@[ IoU=50	&	| area=  all 	&	| maxDets=100 ]	&	40.2	&	39.8	&	15.7	&	0.185	&	0.606	&	0.296	&	0.302	&	0.705	\\
& Avg. Prec. (AP) 	&	@[ IoU=75	&	| area=  all 	&	| maxDets=100 ]	&	37.1	&	37.4	&	13.2	&	0.153	&	0.469	&	0.174	&	0.196	&	0.685	\\
& Avg. Prec. (AP) 	&	@[ IoU=50:95	&	| area= small 	&	| maxDets=100 ]	&	8.5	&	9.4	&	-1	&	0.001	&	0.221	&	0.038	&	0.08	&	-1	\\
& Avg. Prec. (AP) 	&	@[ IoU=50:95	&	| area=medium 	&	| maxDets=100 ] 	&	39.8	&	39.5	&	0.001	&	0.045	&	0.488	&	0.183	&	0.208	&	-1	\\
& Avg. Prec. (AP) 	&	@[ IoU=50:95	&	| area= large  	&	| maxDets=100 ]	&	55.2	&	54.8	&	14.4	&	0.322	&	0.587	&	0.315	&	0.3	&	0.654	\\
& Avg. Rec.  (AR)	&	@[ IoU=50:95	&	| area=  all  	&	| maxDets= 1 ]	&	60.4	&	60.6	&	36.7	&	0.454	&	0.357	&	0.192	&	0.221	&	0.765	\\
& Avg. Rec.  (AR)	&	@[ IoU=50:95	&	| area=  all 	&	| maxDets= 10 ]	&	64.1	&	66.1	&	41.6	&	0.526	&	0.566	&	0.285	&	0.355	&	0.783	\\
& Avg. Rec.  (AR)	&	@[ IoU=50:95	&	| area=  all  	&	| maxDets=100 ]	&	64.3	&	66.2	&	41.7	&	0.527	&	0.594	&	0.293	&	0.374	&	0.783	\\
& Avg. Rec.  (AR)	&	@[ IoU=50:95	&	| area= small  	&	| maxDets=100 ]	&	34.2	&	38.8	&	-1	&	0.188	&	0.367	&	0.055	&	0.17	&	-1	\\
& Avg. Rec.  (AR)	&	@[ IoU=50:96	&	| area=medium 	&	| maxDets=100 ]	&	76.8	&	78.3	&	23.5	&	0.537	&	0.655	&	0.311	&	0.386	&	-1	\\
& Avg. Rec.  (AR)	&	@[ IoU=50:97	&	| area= large 	&	| maxDets=100 ]	&	85.8	&	87.1	&	41.8	&	0.758	&	0.759	&	0.535	&	0.575	&	0.783	\\
\hline																							

\multirow{12}{*} {\textbf{\rotatebox[origin=c]{90}{SSD}}} & Avg. Prec. (AP) 	&	@[ IoU=50:95 	&	| area=  all 	&	| maxDets=100 ]	&	27.4	&	26	&	10	&	0.134	&	0.305	&	0.151	&	0.121	&	0.523	\\
& Avg. Prec. (AP) 	&	@[ IoU=50	&	| area=  all 	&	| maxDets=100 ]	&	36.7	&	33.4	&	14.2	&	0.189	&	0.486	&	0.266	&	0.222	&	0.673	\\
& Avg. Prec. (AP) 	&	@[ IoU=75	&	| area=  all 	&	| maxDets=100 ]	&	32.3	&	30.4	&	11.2	&	0.149	&	0.329	&	0.152	&	0.119	&	0.628	\\
& Avg. Prec. (AP) 	&	@[ IoU=50:95	&	| area= small 	&	| maxDets=100 ]	&	7	&	4.6	&	-1	&	0.001	&	0.098	&	0.02	&	0.04	&	-1	\\
& Avg. Prec. (AP) 	&	@[ IoU=50:95	&	| area=medium 	&	| maxDets=100 ] 	&	31.4	&	29.3	&	0	&	0.029	&	0.357	&	0.152	&	0.126	&	-1	\\
& Avg. Prec. (AP) 	&	@[ IoU=50:95	&	| area= large  	&	| maxDets=100 ]	&	45.1	&	45.2	&	10.8	&	0.257	&	0.484	&	0.309	&	0.225	&	0.523	\\
& Avg. Rec.  (AR)	&	@[ IoU=50:95	&	| area=  all  	&	| maxDets= 1 ]	&	45.8	&	43.1	&	20	&	0.288	&	0.273	&	0.172	&	0.156	&	0.581	\\
& Avg. Rec.  (AR)	&	@[ IoU=50:95	&	| area=  all 	&	| maxDets= 10 ]	&	47.3	&	44.6	&	21.5	&	0.317	&	0.407	&	0.246	&	0.236	&	0.594	\\
& Avg. Rec.  (AR)	&	@[ IoU=50:95	&	| area=  all  	&	| maxDets=100 ]	&	47.5	&	44.7	&	21.7	&	0.321	&	0.429	&	0.259	&	0.255	&	0.594	\\
& Avg. Rec.  (AR)	&	@[ IoU=50:95	&	| area= small  	&	| maxDets=100 ]	&	16.4	&	11.5	&	-1	&	0.08	&	0.152	&	0.026	&	0.076	&	-1	\\
& Avg. Rec.  (AR)	&	@[ IoU=50:96	&	| area=medium 	&	| maxDets=100 ]	&	57.7	&	53.8	&	5.7	&	0.243	&	0.503	&	0.269	&	0.265	&	-1	\\
& Avg. Rec.  (AR)	&	@[ IoU=50:97	&	| area= large 	&	| maxDets=100 ]	&	71.3	&	71.1	&	21.8	&	0.496	&	0.634	&	0.5	&	0.427	&	0.594	\\ 

\hline
\end{tabular}
\end{scriptsize}
\end{center}
\caption{Complete results of the invariance analysis experiments.}
\label{tab:invariance_all}
\end{table*}

\afterpage{ 
\clearpage 
\begin{table*}[htbp]
\begin{scriptsize}
\renewcommand{\tabcolsep}{1.7pt}
\renewcommand\arraystretch{.9}
\vspace{-50pt}
\hspace{-50pt}
\begin{tabular}{l|lllllll|lllllll|lllllll|lllllll|} 

 \textbf{Category} &  \multicolumn{7}{c|} {\textbf{FasterRCNN}} & \multicolumn{7}{c|} {\textbf{RetinaNet}} &  \multicolumn{7}{c|} {\textbf{FCOS}} & \multicolumn{7}{c} {\textbf{SSD}} \\
 \hline\hline
 
& \rotatebox[origin=l]{90}{whiteBG}  & \rotatebox[origin=l]{90}{noiseBG} & \rotatebox[origin=l]{90}{crop\_res.} & \rotatebox[origin=l]{90}{crop} & \rotatebox[origin=l]{90}{objs.\_only} & \rotatebox[origin=l]{90}{blur} & \rotatebox[origin=l]{90}{vert.\_flip} & \rotatebox[origin=l]{90}{whiteBG}  & \rotatebox[origin=l]{90}{noiseBG} & \rotatebox[origin=l]{90}{crop\_res.} & \rotatebox[origin=l]{90}{crop} & \rotatebox[origin=l]{90}{objs.\_only} & \rotatebox[origin=l]{90}{blur} & \rotatebox[origin=l]{90}{vert.\_flip} & \rotatebox[origin=l]{90}{whiteBG}  & \rotatebox[origin=l]{90}{noiseBG} & \rotatebox[origin=l]{90}{crop\_res.} & \rotatebox[origin=l]{90}{crop} & \rotatebox[origin=l]{90}{objs.\_only} & \rotatebox[origin=l]{90}{blur} & \rotatebox[origin=l]{90}{vert.\_flip} & \rotatebox[origin=l]{90}{whiteBG}  & \rotatebox[origin=l]{90}{noiseBG} & \rotatebox[origin=l]{90}{crop\_res.} & \rotatebox[origin=l]{90}{crop} & \rotatebox[origin=l]{90}{objs.\_only} & \rotatebox[origin=l]{90}{blur} & \rotatebox[origin=l]{90}{vert.\_flip} \\
\hline\hline

person	&	51.7	&	48	&	21.6	&	16	&	50.7	&	31.2	&	17.7	&	51.1	&	51.3	&	26.9	&	34.8	&	51.4	&	34.4	&	20.7	&	56.1	&	53.7	&	20.6	&	31.9	&	56.4	&	34.5	&	21.6	&	41.8	&	40.9	&	18.5	&	31.8	&	40.5	&	26.2	&	13.5	\\
bicycle	&	26.5	&	25.1	&	5.6	&	4.4	&	24.9	&	8.3	&	6.8	&	28.9	&	26.2	&	8.2	&	11.2	&	27.2	&	10.9	&	6.3	&	27.8	&	28.3	&	4.5	&	7.1	&	29.8	&	10.3	&	4.9	&	24.1	&	20.3	&	5.7	&	7.7	&	22.3	&	8.4	&	3	\\
car	&	28.5	&	23.9	&	1.6	&	1.4	&	38.2	&	16.1	&	2.6	&	36.8	&	27.6	&	2.7	&	5.4	&	39.2	&	17.9	&	3.2	&	36.4	&	29.2	&	4	&	7.4	&	43	&	18	&	3.5	&	28	&	24.9	&	1.5	&	3.5	&	28.8	&	13	&	1.4	\\
motorcycle	&	47.1	&	44.4	&	17.7	&	16.9	&	38.2	&	20.1	&	10.9	&	47.2	&	45.6	&	23.3	&	27.8	&	42.3	&	22.4	&	12.7	&	51	&	49.8	&	20	&	24.3	&	41.6	&	20	&	10.9	&	40.6	&	36.9	&	22.1	&	29.4	&	32.2	&	18.9	&	8.8	\\
airplane	&	66.1	&	60.7	&	31.8	&	24.7	&	53.4	&	30.7	&	37.9	&	65.3	&	56.9	&	36.3	&	38.4	&	59.9	&	40.9	&	50.1	&	67.3	&	61.4	&	38.7	&	42	&	66.3	&	45.2	&	52.6	&	55.8	&	51.6	&	41.2	&	46.8	&	49.4	&	31.8	&	35.8	\\
bus	&	62.1	&	60.2	&	36.4	&	23.9	&	57.1	&	36.6	&	20.7	&	64.5	&	59.7	&	41.8	&	43.9	&	64.6	&	43.4	&	29.7	&	70.3	&	64.2	&	41	&	43.8	&	69.6	&	45	&	28.2	&	59.8	&	55.8	&	44.4	&	50	&	58.7	&	38.4	&	20.1	\\
train	&	67.9	&	70.7	&	47.1	&	34.6	&	54.8	&	33.6	&	24.6	&	67.3	&	66	&	49.3	&	51.6	&	63.3	&	45.3	&	33.7	&	70.8	&	68.1	&	49.1	&	56.5	&	69.3	&	45.2	&	31	&	64.1	&	63.2	&	55.1	&	58.4	&	60	&	38.9	&	23.2	\\
truck	&	40.3	&	37.7	&	9.2	&	5.8	&	29.3	&	11.1	&	3.5	&	39.1	&	33.6	&	15	&	18.6	&	34.7	&	18.6	&	4.9	&	40.9	&	36.4	&	13.3	&	16.4	&	39	&	16.8	&	4.9	&	35.7	&	33.5	&	13.9	&	18.3	&	29.3	&	13.9	&	3.5	\\
boat	&	25.1	&	23.7	&	5	&	2	&	19.6	&	6.4	&	2.7	&	26.4	&	21.6	&	4.8	&	6.3	&	20.4	&	8.7	&	1.9	&	29	&	22.7	&	5.9	&	6.6	&	27.4	&	7.7	&	1.5	&	18	&	16.8	&	3	&	4	&	14.7	&	4.7	&	1.4	\\
traffic light	&	43.5	&	27.6	&	0.1	&	0	&	41.2	&	8.4	&	19.9	&	40	&	26.1	&	0.6	&	1.1	&	44.5	&	10	&	18.6	&	36.6	&	26.7	&	0.1	&	0.4	&	48.4	&	10	&	19.2	&	21.6	&	11.8	&	0.1	&	0.1	&	25.5	&	4	&	9.8	\\
fire hydrant	&	61.8	&	61.6	&	27.9	&	21.5	&	58.9	&	39.8	&	32.7	&	64.4	&	63.7	&	34.6	&	45.2	&	63.3	&	45.4	&	41.2	&	63.2	&	61.7	&	27.7	&	41.1	&	67.3	&	48.5	&	40.9	&	53.9	&	56.2	&	26.9	&	38	&	50.3	&	36.7	&	30	\\
stop sign	&	59.8	&	61.5	&	42.1	&	37	&	61.5	&	45.8	&	58.7	&	58.6	&	58.9	&	45.2	&	49.8	&	60	&	49.3	&	56.1	&	58.3	&	61.1	&	38.4	&	45.3	&	63.8	&	48	&	56.9	&	53.2	&	55	&	36.8	&	50	&	51.8	&	40.2	&	49.3	\\
parking meter	&	47.2	&	46.7	&	12.1	&	7.3	&	40.8	&	26.9	&	13.3	&	53	&	54.5	&	17.1	&	27.4	&	52.2	&	25	&	15.1	&	52.4	&	52.1	&	13.2	&	21.7	&	54.7	&	28	&	14.2	&	37.8	&	42.8	&	8.7	&	14.3	&	36.4	&	19.6	&	14	\\
bench	&	25.7	&	24.6	&	7.8	&	5.8	&	20.6	&	9.5	&	2	&	23.5	&	21.2	&	10.2	&	13.4	&	22.5	&	12.7	&	3	&	24.8	&	23.1	&	8.7	&	8.2	&	23	&	10.9	&	2	&	20.1	&	18.6	&	6.8	&	9.4	&	17.3	&	8.5	&	2.6	\\
bird	&	24.4	&	23.4	&	4.9	&	2.7	&	28.7	&	10.6	&	9.1	&	29	&	26.7	&	6.6	&	7.8	&	30.8	&	13.8	&	13.4	&	30.5	&	27.3	&	4.7	&	5.1	&	37.5	&	13.8	&	10.6	&	18.6	&	17.1	&	5.6	&	8.2	&	22	&	8	&	6.2	\\
cat	&	47.7	&	51.8	&	25.2	&	19.4	&	56.2	&	27.9	&	33.1	&	56.4	&	59.3	&	33.8	&	37.3	&	69.8	&	41.5	&	46.6	&	57.4	&	35.2	&	32.2	&	30.4	&	72.8	&	43.9	&	52.4	&	42.2	&	45.8	&	9.5	&	30.9	&	55.6	&	32.8	&	32.4	\\
dog	&	44.5	&	48.2	&	18.5	&	9.8	&	50.9	&	28.7	&	19.8	&	56.1	&	60.5	&	29.1	&	34	&	62.9	&	43.1	&	31.5	&	56.6	&	52.2	&	16.5	&	21.5	&	64.3	&	40.2	&	33.3	&	38.1	&	35.6	&	16.5	&	24.1	&	51	&	27.2	&	15	\\
horse	&	57.4	&	56.4	&	25.4	&	18.7	&	49.4	&	26.9	&	9.6	&	59.3	&	58.7	&	32.8	&	40.1	&	55.1	&	35.2	&	15.3	&	58.1	&	60	&	28.8	&	33.1	&	58.2	&	34.8	&	12.2	&	49	&	48.2	&	25.7	&	34.3	&	47.2	&	22.4	&	7.1	\\
sheep	&	37.6	&	36.6	&	8.2	&	4.9	&	40.1	&	19.7	&	2.6	&	43.5	&	44.3	&	14.5	&	21.5	&	45	&	24.6	&	3.6	&	44.4	&	44.4	&	13.6	&	21.4	&	51.5	&	25.6	&	4.8	&	30.1	&	25.7	&	5.7	&	12.6	&	33.5	&	17.3	&	1	\\
cow	&	41.8	&	39.9	&	10.9	&	10	&	45.7	&	19	&	8.1	&	47.4	&	46.2	&	16.8	&	23	&	51.8	&	25	&	10.3	&	47	&	47.5	&	16.1	&	22.7	&	58.9	&	26.8	&	11.6	&	36.6	&	32.2	&	10.2	&	16.7	&	38.9	&	17.4	&	6.5	\\
elephant	&	63.2	&	61.6	&	30.3	&	28.5	&	54.9	&	31.1	&	11.8	&	66.4	&	66.9	&	40.1	&	47.4	&	61.1	&	38.6	&	16.8	&	67.4	&	66.8	&	35.5	&	42.5	&	66	&	38.3	&	16.1	&	52.2	&	55.8	&	31.6	&	39.3	&	51.9	&	27.2	&	12.1	\\
bear	&	67.8	&	68.6	&	48.8	&	42.4	&	60.5	&	30.4	&	14.3	&	74.4	&	73.9	&	54.8	&	58.4	&	70	&	47.5	&	25	&	75	&	65.2	&	60.5	&	61.3	&	73.7	&	46.2	&	29.3	&	64.8	&	57.6	&	46.3	&	56.6	&	61.5	&	32.1	&	11.8	\\
zebra	&	74.8	&	73.4	&	47.2	&	40.4	&	59	&	30.4	&	38	&	75.3	&	75.5	&	52.5	&	55.7	&	62.6	&	41.8	&	48.3	&	75.5	&	73.2	&	47.1	&	58.1	&	67.3	&	37.4	&	47.7	&	68.6	&	68.1	&	40.9	&	52.8	&	55.5	&	29.9	&	34.9	\\
giraffe	&	70.2	&	70.6	&	42.8	&	31.3	&	62.9	&	31.8	&	37	&	72.1	&	72.8	&	49.5	&	58.6	&	66.8	&	41.3	&	41.7	&	74.1	&	76.2	&	45.6	&	58.1	&	72.2	&	37.9	&	39.5	&	64	&	63.8	&	42.5	&	60.1	&	56.6	&	27.4	&	28.4	\\
backpack	&	4.4	&	3.7	&	0	&	0	&	14.3	&	2.1	&	2.6	&	7.9	&	6.9	&	0.3	&	0.5	&	16.6	&	3	&	3.1	&	8.8	&	9.3	&	0.3	&	0.7	&	16.6	&	3.6	&	2.4	&	7.6	&	6.2	&	0	&	0.1	&	7.8	&	1.8	&	2	\\
umbrella	&	28.8	&	29.7	&	8.8	&	5	&	34.3	&	18.4	&	6.1	&	29.9	&	31.2	&	11.6	&	12.6	&	39.5	&	23.6	&	12.7	&	29.9	&	28.4	&	8.3	&	9.3	&	42.4	&	23.3	&	11.5	&	32.6	&	29.7	&	7.3	&	8.2	&	31.1	&	16.6	&	6.4	\\
handbag	&	1.7	&	4	&	0	&	0	&	11.7	&	2.1	&	0.3	&	4.3	&	5.6	&	0.1	&	0.2	&	15.6	&	4.1	&	0.9	&	3.9	&	7.3	&	0.1	&	0.4	&	18.4	&	4.9	&	0.7	&	3.1	&	4.1	&	0	&	0	&	7.2	&	1.1	&	0.5	\\
tie	&	4.6	&	7.4	&	0.2	&	0.1	&	28.7	&	14.5	&	5.9	&	4	&	4.1	&	0.4	&	0.8	&	29.6	&	18	&	5.9	&	5.6	&	6.6	&	0.6	&	1	&	33.5	&	18.1	&	5.4	&	4.8	&	5.4	&	0	&	0	&	17.8	&	10.2	&	3.9	\\
suitcase	&	25.5	&	27.6	&	4.5	&	2.1	&	31.4	&	7.3	&	10.9	&	27.9	&	27	&	8.6	&	11	&	36.5	&	12.9	&	16.3	&	31.8	&	26.6	&	6	&	5.5	&	43.4	&	10.5	&	16.5	&	22.1	&	22	&	2.6	&	4.2	&	22.5	&	6.2	&	8.9	\\
frisbee	&	12.7	&	17.1	&	1.2	&	0.4	&	49.4	&	26.5	&	44.1	&	17	&	20.9	&	1.1	&	2.6	&	53.9	&	34.3	&	49.2	&	19.8	&	28.2	&	0.7	&	1.7	&	58.1	&	32	&	51.1	&	11.5	&	17.6	&	0.1	&	0.2	&	33.8	&	15.8	&	27	\\
skis	&	12.9	&	8.6	&	0.5	&	0.1	&	22.1	&	8.1	&	7.2	&	9.6	&	5.6	&	0.3	&	0.5	&	18.7	&	8.4	&	6.9	&	12.6	&	8.3	&	0.4	&	0.5	&	24	&	8.1	&	8.8	&	7.2	&	6.6	&	0.1	&	0.2	&	14	&	5.5	&	4	\\
snowboard	&	13.3	&	11.6	&	0.2	&	0.1	&	26.3	&	9.4	&	9.2	&	8.3	&	7.7	&	0.2	&	0.2	&	21.3	&	9.4	&	11.1	&	11.7	&	7.8	&	0.2	&	0.2	&	31	&	12.4	&	16.4	&	7.4	&	3.7	&	0.1	&	0.1	&	20.6	&	7.3	&	4.2	\\
sports ball	&	16.2	&	25	&	0	&	0	&	34.6	&	14.4	&	22.6	&	21.9	&	23.7	&	0	&	0.1	&	36.6	&	15.4	&	28.3	&	18.5	&	30.5	&	0	&	0.1	&	42	&	15.8	&	29.5	&	7.6	&	16.8	&	0	&	0	&	18.7	&	9.8	&	10.6	\\
kite	&	7.8	&	14.4	&	2.8	&	1.4	&	28.2	&	14.1	&	24.7	&	11	&	14.8	&	3.2	&	3.8	&	30.2	&	18.5	&	26.6	&	8.8	&	13.9	&	2.2	&	2.4	&	34.9	&	17.3	&	26.8	&	21.9	&	13.8	&	1.1	&	1.1	&	23.5	&	10.3	&	16.7	\\
baseball bat	&	6.3	&	9.4	&	0.6	&	0.2	&	19.3	&	8.3	&	11.1	&	10.1	&	11.2	&	0.8	&	1.3	&	23.1	&	9.5	&	10.1	&	9.5	&	12.8	&	0.7	&	1.7	&	26.8	&	9.7	&	12.8	&	8.6	&	8.1	&	0.1	&	0.1	&	14.4	&	7.9	&	5.4	\\
baseball glove	&	1.1	&	1.2	&	0	&	0	&	31.8	&	12	&	14.6	&	1.7	&	1.8	&	0	&	0.2	&	34.4	&	16.6	&	19	&	2	&	3.7	&	0	&	0.2	&	35.5	&	14.6	&	19.9	&	1	&	1.8	&	0	&	0	&	20.2	&	7.7	&	8	\\
skateboard	&	25.5	&	26.6	&	1.7	&	0.8	&	44.9	&	18.7	&	15.5	&	26.4	&	28.3	&	2.3	&	4.3	&	50.3	&	22.7	&	22.9	&	26.3	&	28.5	&	2.4	&	3.4	&	55.5	&	23.9	&	25.9	&	15.4	&	13.8	&	0.4	&	0.8	&	35.4	&	14.7	&	10.6	\\
surfboard	&	21.4	&	25.4	&	3.9	&	1.3	&	28	&	13.4	&	20.2	&	21.6	&	25.1	&	3.9	&	5.4	&	29.8	&	17.3	&	19.4	&	23.4	&	22.8	&	4.4	&	4.8	&	34.1	&	17.9	&	18.7	&	16.6	&	19.1	&	0.8	&	0.7	&	21.5	&	12	&	14.2	\\
tennis racket	&	27.8	&	25.1	&	1.3	&	0.4	&	41	&	19.4	&	34.4	&	30.9	&	29.8	&	3.6	&	5.9	&	45.8	&	25.1	&	38.6	&	31.1	&	31.6	&	2.9	&	4.5	&	50.2	&	23.3	&	40.6	&	20.9	&	18.1	&	0.5	&	2.3	&	31.5	&	12.5	&	25.5	\\
bottle	&	20.2	&	21.2	&	0	&	0.1	&	34.8	&	9.9	&	6.3	&	19	&	18.3	&	0.2	&	0.7	&	37.1	&	11.6	&	10.7	&	19.8	&	21.1	&	0.2	&	1	&	40.7	&	11.3	&	9.5	&	13	&	9.8	&	0	&	0	&	21	&	5.3	&	3.1	\\
wine glass	&	15.3	&	11.5	&	0.1	&	0.2	&	31.7	&	8.2	&	5.1	&	17.2	&	15.7	&	0.3	&	0.9	&	34.2	&	8.9	&	8.5	&	19.2	&	17	&	0.4	&	0.7	&	36.7	&	8	&	8.9	&	11.8	&	7.1	&	0.1	&	0.4	&	21	&	5.5	&	2.1	\\
cup	&	14.6	&	17.7	&	0.3	&	0.3	&	38.3	&	12.5	&	9.4	&	12.8	&	13.8	&	0.6	&	2.2	&	41	&	16	&	11.8	&	16.5	&	20.6	&	0.6	&	1.8	&	43.3	&	16.1	&	12.4	&	12.2	&	12.2	&	0.1	&	0.5	&	29.3	&	10.4	&	4.7	\\
fork	&	10.5	&	12	&	0.2	&	0.1	&	24	&	9.7	&	15.4	&	13	&	12.9	&	0.4	&	0.8	&	27.9	&	11.4	&	19.2	&	13.9	&	14.2	&	0.3	&	0.6	&	33.2	&	11.4	&	24.2	&	7.9	&	8	&	0.1	&	0.1	&	14.8	&	6.6	&	8.6	\\
knife	&	3.7	&	5.1	&	0.3	&	0.1	&	11.6	&	3.9	&	5.4	&	4.6	&	4.6	&	0.3	&	0.4	&	13.6	&	5.2	&	9	&	5.4	&	7	&	0.2	&	0.3	&	15	&	4.9	&	9.1	&	2.6	&	3.5	&	0	&	0.1	&	6.2	&	2.4	&	3	\\
spoon	&	2.5	&	2.4	&	0	&	0	&	11.7	&	2.7	&	4.4	&	3.8	&	3.6	&	0.1	&	0.1	&	14.1	&	4.9	&	6.7	&	3.6	&	3.9	&	0.1	&	0.1	&	15.4	&	4.3	&	7.9	&	1.6	&	1	&	0	&	0	&	5.6	&	2.7	&	3	\\
bowl	&	23	&	25.9	&	3.2	&	3	&	41.1	&	19.7	&	11.4	&	20.9	&	23.1	&	5	&	7.3	&	40.8	&	21.5	&	13.4	&	22.5	&	28.5	&	4.1	&	5.9	&	42.4	&	20.6	&	12.6	&	19.3	&	18.1	&	3.4	&	4.9	&	32	&	15.8	&	10.5	\\
banana	&	35.5	&	37.8	&	6	&	5.7	&	22.2	&	9.1	&	14.8	&	37.5	&	38.4	&	8.5	&	13.7	&	26.4	&	12.7	&	17.2	&	44.1	&	45.4	&	4.5	&	11.1	&	24.7	&	11	&	15.7	&	28.9	&	29.4	&	5	&	10	&	16.9	&	7.4	&	11.6	\\
apple	&	27.5	&	26.2	&	1.5	&	1.1	&	21	&	10	&	10.7	&	28.2	&	27	&	3.7	&	5.7	&	22.7	&	10.1	&	14.6	&	30.5	&	34.2	&	1.5	&	2.6	&	22.9	&	9.2	&	15	&	21	&	18.3	&	0.2	&	0.7	&	16.4	&	6.1	&	7.8	\\
sandwich	&	24.7	&	26	&	7.1	&	4.4	&	30.9	&	20.6	&	15	&	31.7	&	32.6	&	11.4	&	11.5	&	35.2	&	21.8	&	14.7	&	33.9	&	38	&	11.2	&	10.5	&	37	&	18.4	&	13.8	&	25	&	18	&	12.4	&	12.1	&	31.8	&	17.8	&	10.3	\\
orange	&	29.8	&	26.2	&	1.7	&	3.2	&	28.6	&	18.3	&	23.8	&	31.4	&	27.6	&	7.9	&	14.1	&	32.6	&	22.7	&	25.6	&	38.4	&	35.9	&	4.5	&	9.8	&	33.5	&	24.5	&	24.5	&	23.9	&	14.5	&	1.3	&	3.7	&	24.6	&	14.9	&	19.8	\\
broccoli	&	23.8	&	27.3	&	1.1	&	6.4	&	23.2	&	9.9	&	18.6	&	31.6	&	29.4	&	2.5	&	6.7	&	25.3	&	11.6	&	18.9	&	35.8	&	33.5	&	1.3	&	3.5	&	24.8	&	13.1	&	20.5	&	23.7	&	8.4	&	0.6	&	4.3	&	19.4	&	8.6	&	15	\\
carrot	&	12.2	&	12.7	&	0.2	&	0.3	&	17.9	&	10	&	17.6	&	12.1	&	10.2	&	1.2	&	3.7	&	21.1	&	11.3	&	18	&	14.4	&	14.8	&	0.6	&	2.3	&	21	&	14.6	&	18.3	&	7.2	&	4	&	0.1	&	0.1	&	13	&	6.5	&	10.6	\\
hot dog	&	15.1	&	20.7	&	4	&	3.3	&	26.2	&	10.4	&	15.9	&	18.5	&	22.7	&	8.5	&	10.5	&	32.9	&	18.7	&	25.1	&	22.3	&	28.2	&	6.7	&	8.4	&	36.7	&	20	&	23.9	&	17.8	&	22.5	&	6.7	&	9.4	&	24.6	&	10.8	&	11.7	\\
pizza	&	36.6	&	40.5	&	17.3	&	14.6	&	46.4	&	34.1	&	37.4	&	41.8	&	43.5	&	22.9	&	26.9	&	50.6	&	37.7	&	40.3	&	51	&	52.8	&	31.9	&	37.3	&	55.4	&	39.7	&	43.8	&	37.5	&	34.3	&	21	&	24.9	&	44	&	31.5	&	33.6	\\
donut	&	24.3	&	24.8	&	1.4	&	2.3	&	43.2	&	18.9	&	24.9	&	23.2	&	26.9	&	4.4	&	4.8	&	46.6	&	22.7	&	28.6	&	27.4	&	35.8	&	2.7	&	3.4	&	50.5	&	26.1	&	30.4	&	21.5	&	20.3	&	2.9	&	3	&	35.6	&	18.7	&	19.3	\\
cake	&	19	&	20.9	&	4.1	&	3.1	&	30.8	&	12.8	&	8.1	&	17.1	&	20.7	&	6	&	7.8	&	35.8	&	15.9	&	7.4	&	18	&	22.5	&	4.8	&	6.4	&	36.7	&	13.8	&	7.3	&	15.9	&	10.7	&	5	&	5.7	&	26	&	11.1	&	5.2	\\
chair	&	19.9	&	21.9	&	0.8	&	0.5	&	25.3	&	8.5	&	2.4	&	21.9	&	22.6	&	1.6	&	3.8	&	28.1	&	11.1	&	3.3	&	25.1	&	25.8	&	1.2	&	3.4	&	32.1	&	10.8	&	3.4	&	17.9	&	15.8	&	0.4	&	0.5	&	19.6	&	7	&	1.3	\\
couch	&	40.4	&	41.5	&	17.6	&	12.2	&	35.8	&	14.6	&	2.6	&	43.1	&	44.6	&	25.8	&	29.9	&	41.1	&	22.8	&	5.1	&	43.3	&	42.7	&	19.7	&	19.9	&	44.5	&	19.8	&	5	&	36.8	&	34.9	&	12	&	15.2	&	36.3	&	13.2	&	2.4	\\
potted plant	&	35.5	&	29.5	&	1.8	&	2	&	34.1	&	8.4	&	3.6	&	37.1	&	27.6	&	2.6	&	5.4	&	33.8	&	8.4	&	4.4	&	37.1	&	32.7	&	1.8	&	4	&	37.4	&	8.6	&	4.8	&	29.1	&	20.6	&	2.3	&	3.1	&	25.4	&	4.9	&	2.5	\\
bed	&	51.1	&	54	&	36.3	&	29.8	&	39	&	22.2	&	10.9	&	51.6	&	49.7	&	39.8	&	41.3	&	49.3	&	32.1	&	17.2	&	40.9	&	38.8	&	29.2	&	30.5	&	50.5	&	23.2	&	13.1	&	42.9	&	39.2	&	18.8	&	42.1	&	43.2	&	23.1	&	10	\\
dining table	&	34.2	&	33.2	&	27.7	&	23.2	&	29.6	&	16.5	&	10.1	&	32.8	&	29.9	&	28.5	&	30.2	&	32	&	21.1	&	12.1	&	28.5	&	24.3	&	22.2	&	23.9	&	32.2	&	18.2	&	11.7	&	30.6	&	24.5	&	15.8	&	29.8	&	28.2	&	16.5	&	11.8	\\
toilet	&	64.4	&	68.6	&	33.6	&	22.7	&	56.6	&	24.6	&	23.2	&	67.9	&	70.2	&	41.7	&	46.3	&	66	&	38.8	&	27.5	&	68	&	66.7	&	33	&	36.5	&	70.6	&	35.4	&	24.5	&	58.4	&	59	&	34.5	&	41.1	&	57.1	&	29.9	&	19.9	\\
tv	&	53.1	&	52	&	15.5	&	9	&	54.2	&	35.4	&	28.1	&	54.6	&	51.7	&	24.6	&	28.5	&	60.9	&	41.1	&	30.9	&	57.1	&	48.5	&	13.9	&	12.9	&	62.6	&	39.7	&	32.9	&	49.8	&	49.3	&	8.2	&	5.6	&	50.5	&	33	&	24.8	\\
laptop	&	42.2	&	49	&	24	&	19.2	&	51.8	&	28.5	&	9.6	&	45.5	&	50.4	&	32.8	&	36.5	&	58.9	&	37.9	&	11.6	&	43.4	&	44.8	&	18.6	&	20.2	&	61.6	&	33.4	&	11.1	&	43.9	&	46.9	&	20	&	19.1	&	51.4	&	26.3	&	6.3	\\
mouse	&	22.3	&	31.4	&	0.2	&	0.1	&	43.8	&	20.6	&	25	&	24.8	&	30.2	&	0.3	&	0.4	&	46.3	&	23.6	&	28	&	15.4	&	23.6	&	0.2	&	0.8	&	48.9	&	23.1	&	28.3	&	26.8	&	31	&	0	&	0	&	38	&	15.8	&	17	\\
remote	&	3.9	&	5.8	&	0.1	&	0.1	&	23.4	&	6.1	&	11.9	&	5.1	&	7.3	&	0.1	&	0.3	&	26.8	&	7.8	&	13.8	&	4.2	&	8.9	&	0.2	&	0.6	&	32.3	&	8.2	&	14.8	&	5	&	7.1	&	0	&	0.1	&	12.5	&	4.3	&	6.1	\\
keyboard	&	35.4	&	39.6	&	21.8	&	23.7	&	46.8	&	14.8	&	28.8	&	42.3	&	43.5	&	26.1	&	29.5	&	46.9	&	19.2	&	24.5	&	42.4	&	44.8	&	23.7	&	25.6	&	47.3	&	21.8	&	27.1	&	42.5	&	40.5	&	7.1	&	7.1	&	41.4	&	12.7	&	18.2	\\
cell phone	&	7.9	&	11.1	&	0.2	&	0.1	&	29.6	&	13.3	&	14.5	&	8.7	&	11.8	&	0.6	&	1	&	34.1	&	16.9	&	18.2	&	11.7	&	15.2	&	0.6	&	1.1	&	36	&	14.4	&	16.4	&	10.1	&	14.1	&	0.6	&	0.9	&	23.3	&	13.7	&	12.6	\\
microwave	&	37.6	&	38.2	&	9.7	&	3.4	&	53.3	&	29.7	&	36.6	&	45.9	&	46.8	&	17.5	&	18.2	&	52.9	&	33.1	&	40.2	&	48.8	&	46	&	6.8	&	8.2	&	58.3	&	29.4	&	43.3	&	42.3	&	40.3	&	2.6	&	4.3	&	48.9	&	28.1	&	30	\\
oven	&	47	&	48.2	&	14.9	&	11.7	&	36	&	11.5	&	5	&	49.4	&	50.1	&	21.7	&	25.4	&	41.6	&	15.7	&	9.4	&	51.4	&	49.3	&	14	&	14.5	&	44.8	&	13.1	&	5.9	&	43.2	&	42.4	&	16.1	&	23.2	&	36.3	&	10.5	&	7.8	\\
toaster	&	18	&	19.4	&	3.2	&	0.2	&	31.1	&	6.6	&	1.5	&	10.2	&	2.8	&	2.2	&	2.7	&	27.1	&	1.2	&	3.7	&	34.5	&	35	&	1.3	&	0.6	&	57.5	&	0.7	&	3.9	&	3.3	&	0.4	&	0	&	0	&	12.9	&	0.3	&	0.6	\\
sink	&	28.4	&	34.2	&	7.5	&	4.9	&	30.5	&	12.7	&	10.3	&	28.9	&	34.5	&	7.3	&	10.1	&	32.5	&	15	&	9.9	&	30.4	&	34.5	&	7.4	&	7.8	&	35.8	&	14.4	&	9.2	&	27.5	&	28.3	&	3.9	&	3.3	&	26.9	&	10	&	9.1	\\
refrigerator	&	59.8	&	64.3	&	36.9	&	16.5	&	47.2	&	23.7	&	18.9	&	60.7	&	55.4	&	41.7	&	43.2	&	57	&	30.7	&	28.4	&	59.7	&	60.2	&	36.7	&	35.3	&	61	&	30.2	&	27.5	&	53.3	&	53.2	&	34.6	&	36.1	&	46.4	&	20.3	&	17.5	\\
book	&	4.1	&	7.5	&	0.4	&	0	&	16.7	&	4.3	&	8.6	&	5.6	&	6.4	&	0.2	&	0.3	&	15.9	&	5.2	&	8.7	&	7.6	&	8.4	&	0.4	&	0.8	&	17.7	&	3.9	&	7.8	&	4.6	&	4.2	&	0	&	0	&	8.9	&	2.5	&	5	\\
clock	&	54.9	&	50.3	&	9.2	&	7	&	50.9	&	22.7	&	42.6	&	56.9	&	49.6	&	10.9	&	17.4	&	54.8	&	25.2	&	41.9	&	56.7	&	55.1	&	6.2	&	8.2	&	57.4	&	23.4	&	45.3	&	41.7	&	40.3	&	5.3	&	14.9	&	42.1	&	19.5	&	33.8	\\
vase	&	24.4	&	24.9	&	1	&	0.6	&	33.9	&	12.6	&	4.3	&	24.7	&	24.5	&	2.7	&	3.4	&	36.2	&	14.4	&	3.8	&	28.2	&	27.2	&	1.6	&	3	&	42.6	&	12.6	&	4	&	20.5	&	14.3	&	1.3	&	1.7	&	27.7	&	10.2	&	2.7	\\
scissors	&	26	&	27.7	&	13.1	&	2.6	&	16.3	&	12.8	&	11.8	&	33.1	&	34	&	19.4	&	21.1	&	33.3	&	21.4	&	20.8	&	33	&	28.8	&	12.7	&	10.6	&	34.1	&	19.1	&	20.6	&	25.4	&	25.3	&	6.2	&	5.5	&	23.2	&	13.8	&	15	\\
teddy bear	&	42.8	&	46.6	&	17.8	&	15.8	&	41.9	&	20.8	&	13.5	&	48.6	&	50.6	&	26.6	&	29.5	&	48.3	&	26.5	&	18.6	&	51.7	&	52.6	&	24.5	&	26.9	&	53	&	25	&	20.6	&	38.1	&	38.3	&	21.4	&	24.5	&	36.3	&	19.6	&	11.8	\\
hair drier	&	0.6	&	1.1	&	0	&	0	&	2	&	0.1	&	0.2	&	0.4	&	0.4	&	0	&	0	&	0.3	&	0	&	0.1	&	3.6	&	10.6	&	0	&	0	&	12.8	&	2.1	&	0.5	&	0	&	0	&	0	&	0	&	0	&	3	&	0	\\
toothbrush	&	3	&	3.3	&	0.2	&	0.1	&	13.9	&	7.3	&	7.4	&	5	&	6.3	&	0.6	&	0.8	&	17.5	&	10.3	&	7	&	5.9	&	7.2	&	0.3	&	0.3	&	20.2	&	10.6	&	14.5	&	2.6	&	3	&	0	&	0	&	7.5	&	2.1	&	4.1	\\

\hline
\end{tabular}
\end{scriptsize}
\vspace{-5pt}
\caption{Mean AP breakdown over COCO categories (results of the invariance analysis).}
\label{tab:invariance_category}
\end{table*}
\thispagestyle{empty}
\clearpage 
}

\section{Discussion}

\noindent {\bf Summary of the learned lessons.} Through exhaustive analyses, we found that a) models perform significantly below what is empirically possible, b) the performance gap is larger over small objects, indicating that scale is one of the major problems in object detection, c) the bottleneck in object detection is object recognition, and d) detection models lack generalization in terms of searching the right places, utilizing context, recognition of small objects, and robustness to image transformation. 

\noindent {\bf Recognizing objects in natural scenes and empirical upper bound.} What we essentially did in this paper was to build the best object classifier for objects embedded in natural scenes as opposed to isolated objects in object recognition datasets (\eg ImageNet). Using this object classifier we then approximated the empirical upper bound in average precision. In contrast to prior investigations on the influence of context in object recognition (\eg~\cite{rabinovich2007objects}), we did not find a significant contribution from the surrounding context in recognizing a target object, except in one case which was training and testing on 1.8 context and testing on the same condition over the COCO dataset (See Table~\ref{tab:context}). An evidence corroborating this finding was our experiment in placing objects out of their context. Results showed that models are still able to detect objects (at least large objects used here). Rosenfeld~\etal~\cite{rosenfeld2018elephant} performed a similar experiment, but over a much smaller set of images, and found that models fail to detect out of context objects (\eg an elephant in the room shown in Fig.~\ref{fig:elephant}. Since they did not report the AP of models on these images, a fair and careful comparison is not feasible at this point, which leaves answering this question to future investigations. 

The contradiction between our results and previous investigations on the role of visual context might be due to the fact that we performed these experiments in large scale and over a much larger set of objects. A more systematic investigation of this may bring new insights. For example, it is very likely that context will play a more important role for recognition and detection of small objects or occluded ones. Further, we found that data augmentation using boxes overlapping with a target object did not lead to better classification accuracy. A further investigation of this using extensive data augmentation, external data, other backbones, or other optimization approaches may improve the upper bound slightly, but perhaps not significantly. 


We invite researchers to periodically update the upper bound in detection scores including AP and other recently proposed ones such as LRP~\cite{oksuz2018localization}\footnote{It turns out that this score reduces to the classic AP when there is no localization error, thus the UAP computed here also applies to the LRP.} and probability-based detection quality~\cite{hall2018probability}, especially over upcoming large scale datasets such as OpenImages~\cite{kuznetsova2018open}. The same can also be repeated for other tasks such as semantic segmentation, instance segmentation, image and video captioning\footnote{A work on this already exists~\cite{yao2016empirical}.}, saliency prediction~\cite{bylinskii2016should,hou2017deeply,wang2018revisiting,he2019human}, image generation~\cite{goodfellow2014generative,borji2019pros} and detection of specific objects (\eg faces, pedestrians). An ongoing effort is also addressing the shortcomings of detection scores. One line of research known as panoptic segmentation~\cite{kirillov2019panoptic} is encouraging a new way to evaluate the segmentation models. Perhaps in the future, researchers may abandon predicting bounding boxes in images (and hence getting rid of complications in AP calculation) and focus on the panoptic segmentation task which regards classifying all image pixels (into object and stuff classes). In this sense, object detection is a subset of panoptic segmentation.







%
\noindent {\bf Challenges regarding object detection.}
Object detection models (and also object recognition models) perform very well on input images in which objects do not go under (relatively) drastic variations (\eg face detection or face recognition). Even in the case of faces, models suffer from issues such as low light conditions, highly occluded faces, or tiny ones. Detection of generic objects is very difficult due to several challenges. First, objects can be partially occluded (\eg a dogs behind the couch). As a result, the features extracted at the object location are not powerful enough to classify that object. To harness this, a large number of data points covering those scenarios are needed. Second, object appearance and shape vary significantly from different viewpoints (\eg due to in-depth rotation). An object detector trained on specific viewpoints may fail to detect an object from a novel viewpoint at the test time. Third, similarly, objects may appear large or small due to their distance to the camera or due to their natural scale. Currently, object detectors utilize data augmentation or a scale pyramid (\eg feature pyramid network proposed in~\cite{lin2017feature}) to solve this problem. Nonetheless, as we showed here, the scale problem still stands. Fourth, not all objects are rigid. Some non-rigid objects such as cloths can undergo drastic deformations which makes detecting them very challenging. To make matters even worse, a non-rigid object can be split into several (disconnected) parts. Fifth, in some situations, especially in videos, captured images are occasionally blurred due to the object or camera motion. This is critical to overcome in particular for applications involving moving robots, self-driving cars, or drones.

\begin{figure}[htbp]
\begin{center}
   \includegraphics[width=.8\linewidth, angle=0]{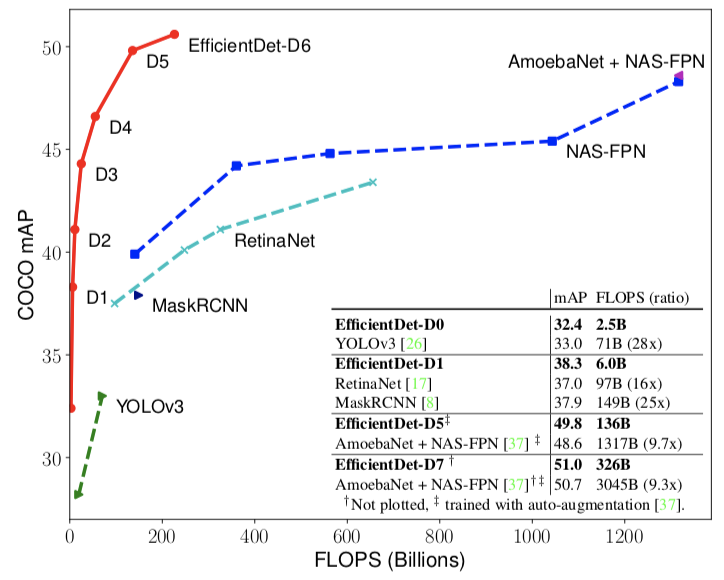}
\end{center}
    \caption{State of the art performance on COCO2017 test-dev dataset according to~\cite{tan2019efficientdet}. EfficientDet achieves much better accuracy with fewer computations than other detectors. It achieves new state-of-the-art 51.0\% COCO mAP with 4x fewer parameters and 9.3x fewer FLOPS. Figure reproduced from~\cite{tan2019efficientdet}.}
\label{fig:efficientDet}
\end{figure}

As we mentioned repeatedly throughout the paper, above problems are not specific to the detection methods and stem from the shortcomings of the recognition component in the detectors. After all, all modern object detectors are based on convolutional neural networks which suffer from the lack of generalization and high demand for a large number of annotated data. Despite these shortcomings, there is still room to improve in object detection, as models perform much lower than the empirical upper bound (calculated in this paper). This is great because it means that we can still significantly boost the object detection performance. The best mAP performance on COCOval2017 dataset is 46.9\% (See Fig.~\ref{fig:coco_fashion_voc}, which is far below the empirical upper bound of 78.2\%. The best mAP over the  COCO2017 test-dev dataset is 51.0\% (See Fig.~\ref{fig:efficientDet}). Since results over the COCO test-dev are usually higher than the results over the COCO validation set, we predict the empirical upper bound to be better over the former set, but most likely there will still be a large gap between models and the empirical upper bound on the test-dev dataset.

\noindent {\bf Error diagnosis and invariance analysis}
We proposed a novel approach to study the errors of object detectors. Our error analysis experiments show that classification errors are much more prevalent than other types of errors and contribute the most to the overall error. This aligns with our argument in the previous section which showed that detection upper bound depends on the recognition accuracy. An alternative approach to evaluate the recognition component of an object detector, is to feed the target boxes to a model and collect its decisions on those boxes. This is, however, cumbersome and needs to be implemented for each model separately\footnote{
A preliminary investigation by feeding GT bounding boxes (at inference time) to FasterRCNN models with ResNet50 backbone and FPN, results in mAP of 73.3\% on COCOval2017.
}, whereas our diagnosis tool is general.


Our new diagnosis tool can be employed to pinpoint weaknesses in other object detection models. Also, error analysis of models for other tasks (\eg object tracking~\cite{alwassel2018diagnosing}) is encouraged. Further, a more systematic investigation of invariance properties of object detectors along with adversarial examples to challenge object detectors can bring new insights into the failures of object detectors and for building better models. In this regard, the \emph{MMDetection} benchmark offers code for analyzing models over transformed images (such as noise, blur, etc). Finally, here we were not concerned with the processing speed of the object detection models. Future work can study empirical upper bound when speed is also a concern.

\noindent {\bf Shall we dismiss object detection?}
As was mentioned in section~\ref{unification}, object detection is tightly related to object recognition and semantic segmentation. In particular, it relates to instance segmentation where the task is to label pixels belonging to individual objects of different classes (\ie distinguishing different cars). In a sense it generalizes many other tasks. A natural question to ask is whether instance segmentation models can outperform object detection models in terms of accuracy and speed. If so, then maybe we should abandon the object detection problem (\ie predicting bounding boxes) and focus on instance segmentation.

To gain insights regarding the above question, we generated bounding boxes from the predicted instance masks by a model, thus creating an object detector. The AP-Box of this object detector is then calculated. Conversely, predicted bounding boxes of an object detector can be considered as instance masks, thus an instance segmentation model. The AP-Mask of this model is then calculated. Performance of five models, all using the R50-FPN backbone, over the MS-COCO val2017 (36781 objects) are shown in Table~\ref{tab:detdismiss}.




The first 4 models in Table~\ref{tab:detdismiss} fall under the category of `detect-then-segment' models whereas the last one, TensorMask~\cite{chen2019tensormask}, performs instance segmentation directly. The latter also does object detection but independently from segmentation. The first (last) two rows in each model show AP-Box (AP-Mask). The second row shows AP when using predicted masks as boxes (\ie circumscribed rectangles) and the fourth row shows applying boxes as masks. Note that all five models generate both boxes and masks.

Results show that AP$\_$Box is higher (about 1 to 2\%) using the predicted bounding boxes than using boxes fitted to segmentation masks. This indicates that predicting bounding boxes directly leads to better accuracy (so far), and faster inference time, indicating that it makes sense to continue working on object detection. According to Table~\ref{tab:detdismiss}, AP$\_$L when using masks as boxs is very close to when using the original predicted boxes. This is perhaps because predicted instance masks over large objects are already very accurate (witnessed by the higher AP$\_$Mask for large objects). Also, applying boxes as masks results in much lower AP-Mask compared to using the original predicted masks (see the fourth rows), since regions from other objects and the image background are also included in the box. \\

\begin{figure}[t]
\begin{center}
   \includegraphics[width=1\linewidth, angle=0]{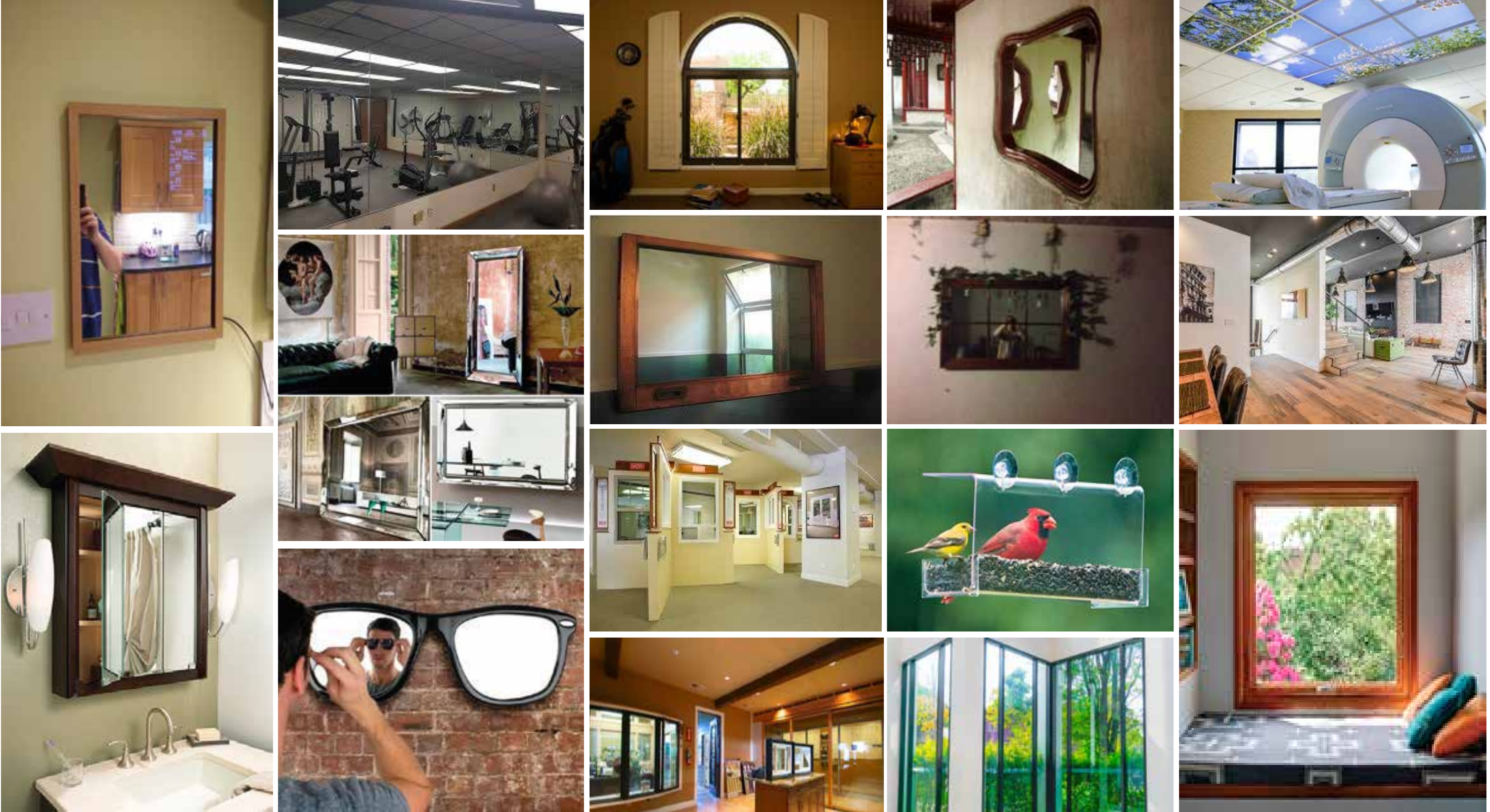}
\end{center}
    \caption{Example mirror or windows/glasses in natural images. Try to find the mirrors. A mirror can be confused with a window or sometimes a painting.}
\label{fig:mirror}
\end{figure}




\begin{table*}
\begin{center}
\begin{small}
\renewcommand{\tabcolsep}{9pt}

\begin{tabular}{l|l|ccc|ccc}
Model & Score & $AP$ & $AP^{.5}$ & $AP^{.75}$ & $AP_S$ & $AP_M$ & $AP_L$  \\

\hline\hline
Mask R-CNN & AP-Box $ \ \ $  [ predicted boxes ] & 
0.373 & 0.590 & 0.402 & 0.219 & 0.409 & 0.481 \\
& AP-Box  $ \ \ $  [ mask as box ] &  0.366 & 0.575 & 0.392 & 0.203 & 0.402 & 0.485 \\
\cline{2-8}
& AP-Mask [ predicted masks ] &  0.342 & 0.559 & 0.362 & 0.158 & 0.369 & 0.501 \\
& AP-Mask [ box as mask ] &  0.123 & 0.352 & 0.066 & 0.075 & 0.132 &  0.166 \\  

\hline
Cascade Mask & AP-Box & 
 0.412 & 0.591 & 0.451 & 0.233 & 0.445 & 0.545 \\
R-CNN & AP-Box  $ \ \ $ [ mask as box ] & 0.394 & 0.580 & 0.428 & 0.212 & 0.426 & 0.536 \\
\cline{2-8}
& AP-Mask &  0.357 & 0.563 & 0.386 & 0.164 & 0.382 & 0.526 \\
& AP-Mask [ box as mask ] &  0.128 & 0.353 & 0.073 & 0.081 & 0.133 & 0.173 \\

\hline
Hybrid Task & AP-Box & 
0.421& 0.608&  0.459&  0.239&  0.455&  0.562 \\ 
Cascade & AP-Box  $ \ \ $ [ mask as box ] & 0.405& 0.598&  0.435&  0.217&  0.438&  0.555 \\
\cline{2-8}
& AP-Mask & 0.373& 0.582&  0.402&  0.174&  0.400&  0.553 \\
& AP-Mask [ box as mask ] & 0.132& 0.367&  0.074&  0.087&  0.139&  0.178 \\

\hline
Cascade Mask & AP-Box & 
0.444& 0.629&  0.484&  0.258&  0.480&  0.595 \\
R-CNN + DCN & AP-Box  $ \ \ $ [ mask as box ] &  0.422& 0.617&  0.459&  0.233&  0.456&  0.578 \\
\cline{2-8}
& AP-Mask & 0.383& 0.600&  0.413&  0.184&  0.409&  0.564 \\
& AP-Mask [ box as mask ] & 0.134& 0.373&  0.077&  0.091&  0.141&  0.179 \\

\hline\hline
TensorMask & AP-Box &  
0.414& 0.607&  0.447&  0.250&  0.441&  0.540 \\
& AP-Box  $ \ \ $ [ mask as box ] & 0.399& 0.589&  0.427&  0.232&  0.438&  0.511 \\
\cline{2-8}
& AP-Mask & 0.358& 0.569&  0.379&  0.175&  0.380&  0.505 \\
& AP-Mask [ box as mask ] & 0.127& 0.358&  0.070&  0.087&  0.135&  0.167

\end{tabular}
\end{small}
\end{center}
\caption{AP results using predicted bounding boxes and instance masks interchangeably.}
\label{tab:detdismiss}
\end{table*}









\begin{figure}[htbp]
\begin{center}
   \includegraphics[width=.8\linewidth, angle=0]{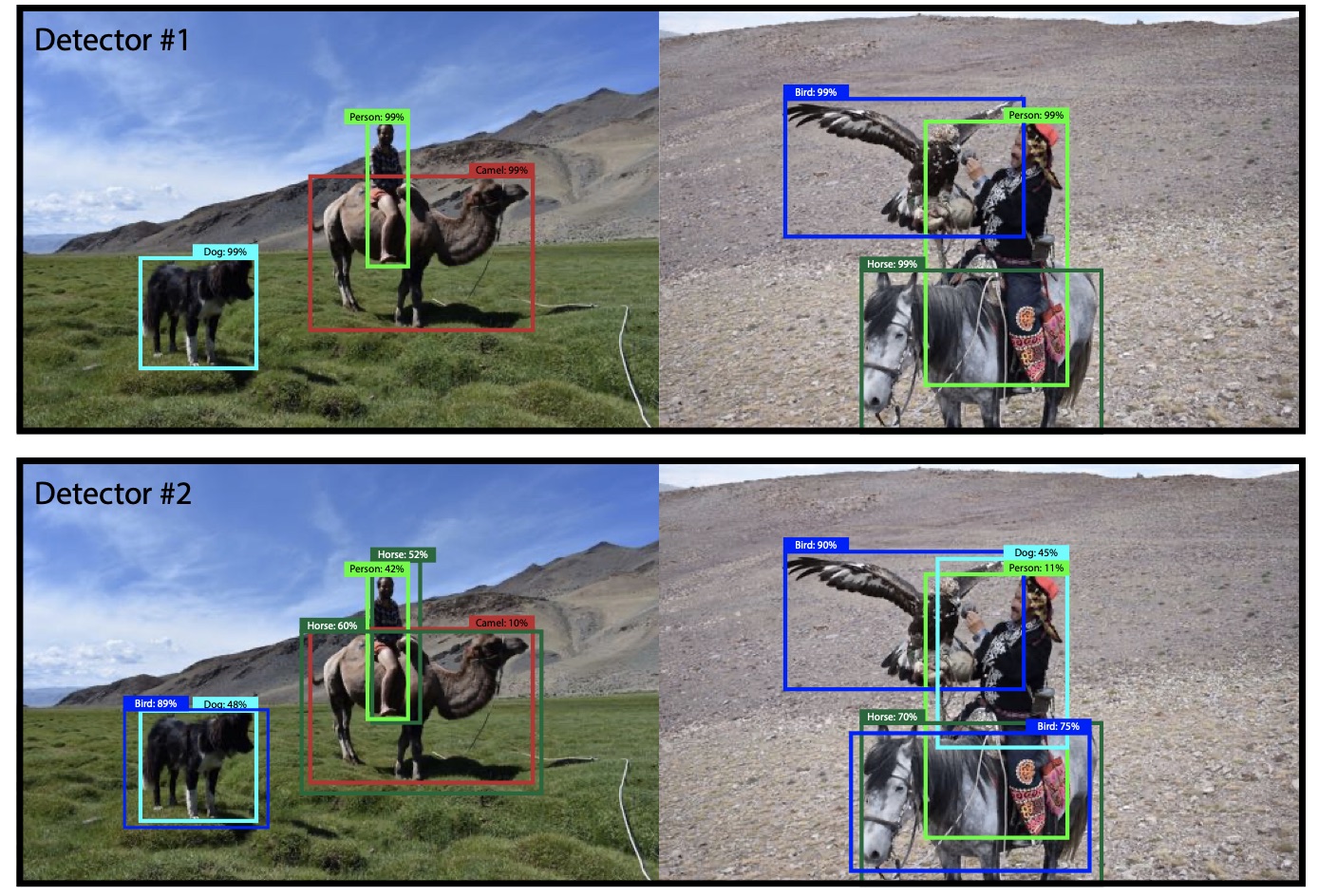}
\end{center}
    \caption{Two hypothetical object detector, one perfect and another with false positives, attain perfect and equal mAP over these two images. Image reproduced from~\cite{redmon2018yolov3}.}
\label{fig:yolo}
\end{figure}

\noindent {\bf Evaluation measures.} Evaluating object detection models has been a matter of debate. mAP is a well-established score but it has several shortcomings. Since mAP is calculated per class, it sometimes generates non-intuitive values. An example is illustrated in Fig.~\ref{fig:yolo} where a detector generating boxes for non-existing objects (\ie false positives) attains perfect mAP. Also, previous research has shown that an object detector with a lot of lower-confidence false positives can win over a detector with comparitively lower false positives (See~\cite{corke2020can}). As was stated in the Introduction section, mAP calculation is complicated. Further, it is unclear how much a small improvement in mAP (say 1\%) will matter in real world applications (see also the discussion in~\cite{redmon2018yolov3}). Eventually, as was discussed above, with the rise of the instance segmentation and its corresponding evaluation measures, the suitability of mAP for object detection demands further discussions.

\noindent {\bf Is localization ignored in our analysis?}
Our main intention here was to assess the power of classification in object detection. Conversely, as a complementary approach, one could take the predictions by a model and ask humans to annotate them. However, this is very cumbersome, whereas our setup is straightforward as it does not need annotations by humans. In fact, our error diagnosis does something similar to the latter. Overall, we aim to understand the power of deep learning in solving object detection. We have not neglected the localization component for the following reasons. First, we hypothesize that our setup gives the empirical upper-bound. So, we had to fix the localization to reach the upper-bound. Any error in localization will only lower the UAP (which will not be upper-bound anymore!). Second, we have investigated how inaccuracy in localization affects UAP (section~\ref{sec:uap}). Third, we have provided a very detailed analysis of the localization error in models (section~\ref{sec:error}).

\noindent {\bf The role of context in object recognition versus object detection.}
In section~\ref{sec:uap}, we found that the surrounding context is not important in classifying the center object (on average). In section~\ref{sec:invariance}, we analyzed context as it is incorporated in current object detection models. Here, context is more important for small objects (compare Table~\ref{tab:invariance_results1} objects\_only vs. Table~\ref{tab:invariance_results2} orig\_img; last columns) since both localization and classification are involved. Overall, to answer how important context is in detecting or recognizing objects depends on how it is utilized.

\noindent {\bf Sampling boxes from the background.}
Here we only considered object boxes to train the classifiers. Since we are looking for the best classifier to only classify objects, including the background class will only lower the accuracy of the classifier. There is no need to include the background since for computing the upper-bound, background regions are already discarded (\ie assuming perfect localization and objectness prediction).


\noindent {\bf Understanding when and why the upper bound fails.} 
In some rare occasions (\eg tunics in Fig.~\ref{fig:fashion}, toaster in Fig.~\ref{fig:coco_detectron}; often small objects), UAP is lower than model AP possibly because our classifier has to elicit a decision for any box, thus it may generate more false positives than a model that misses objects (\ie we do not have misses). This may results in lower precision for some classes for our UAP than a model, but our setup has a higher recall.

\begin{figure}[htbp]
\begin{center}
   \includegraphics[width=.7\linewidth, angle=0]{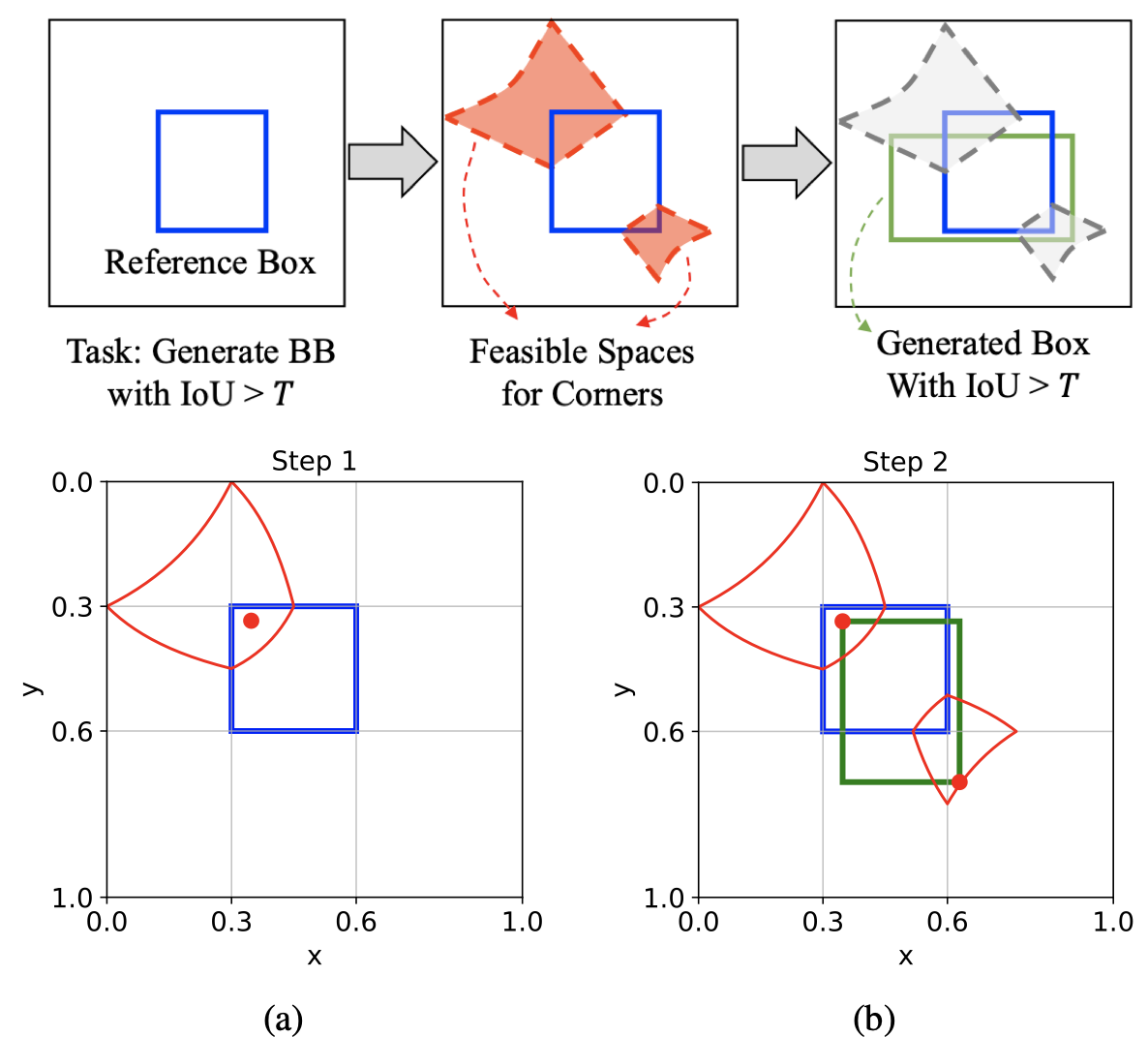}
\end{center}
    \caption{An illustration of the approach taken by~\cite{oksuz2018localization} for sampling boxes at different scales around the ground-truth box (shown in blue). To generate a box (shown in green) with IoU larger than T = 0.5, two points are sampled in the two red polygons. In this example, the green box has the IoU = 0.5071. Image reproduced from~\cite{oksuz2018localization}.}
\label{fig:kemal}
\end{figure}

\noindent {\bf Sampling boxes at different scales.}
Our proposed sampling strategy is efficient at covering the space of translation. However, it does not capture variation in scale. In contrast, two-stage detectors do so through bounding box regression in the first stage. The way we mitigate this is through extensive data augmentation during training the classifier. This classifier is applied to the scaled versions of the boxes surrounding the target box.
A more general box sampling approach than ours has been proposed in parallel by Oksuz~\etal~\cite{oksuz2018localization} which can be used to sample boxes at different scales. See Fig.~\ref{fig:kemal}.


\noindent {\bf One possible reason for lower detection performance on small objects.}
It could be because small objects (\eg a pen) are very small in both train and test sets. It is even hard for humans to recognize small objects out of context. There is not much scale variation in both train and test set in detection datasets over small objects. This is in contrast to object recognition datasets where scale variation is larger. Thus, even data augmentation for these classes may not help detection performance much. To detect or recognize small objects, it might be better to observe them also in large scale. To remedy this problem, one way may be to rely on external data for small objects. 



\noindent {\bf Our error diagnosis compared to Hoeim~\etal.}
The main difference is that instead of removing detections, we fix them. Consider two methods (method 1 and method 2) both generating many mislocalized detections (FPs). For method 1, after correcting the mislocalized detections based on our protocol, many of them recover the misses and become TPs. On the contrary, after correcting the mislocalized detections for method 2, they become redundant to the correct detections (TPs) and hence are considered as FPs. Thus, the mislocalized predictions in the two methods are actually different. Our error analysis is able to discern method 1 and method 2, while the vanilla protocol fails since it removes the mislocalized predictions blindly.

\section{Conclusion and Outlook}
 Modern object detectors are far from perfect despite intensive research in this area and significant progress in deep learning over the last couple of years. This signifies the limitation of deep learning to solve challenging vision tasks suggesting perhaps some fundamental ingredients are missing. Fixing the localization problem leads to the empirical upper-bound but reaching beyond that demands having better object recognition models. 

As it stands robustness to scale still remains the main challenge in object detection. Scale variation (across various objects) is much higher in detection datasets than recognition datasets because objects are captured in their natural habitats. A certain objects might appear in a certain scale most of the time (\ie their natural scale). For example a pen may always appear small compared to other objects in the scene. This makes recognition of such objects, out of their context, very difficult (See Fig.~\ref{fig:collage}). This is less problematic in recognition datasets, such as ImageNet, since objects in those datasets are intentionally selected to be visually recognizable by humans. What all these means is that perhaps we need more data for object detection or we need to resort to external data to improve results on existing detection datasets. Humans are much better in detecting small objects and in exploiting the surrounding context around an object possibly due to the structural differences between the human visual system and CNNs . For instance, human retina consists of a high resolution central region called fovea and a lower resolution peripheral region. By moving the fovea over the scene, our eyes capture finer details of objects, whereas the resolution is fixed in still images fed to CNNs~\cite{borji2012state}.

Lastly, our investigation here shows that we are far from solving the object detection problem. Further, this task can be considered as a litmus test to assess the capacity of deep learning and CNNs for solving vision problems. Existence of adversarial examples against object detection models (\eg~\cite{xie2017adversarial,eykholt2018physical,braunegg2019apricot}) also exacerbates the problem and demonstrates how fragile these models are (and also many other models based on CNNs; See~\cite{goodfellow2014explaining}). Adversarial examples are perhaps a byproduct of the lack of robustness in vision models (\eg \cite{tsipras2018robustness}. 
Designing more powerful architectures (\eg using architecture search techniques~\cite{zoph2016neural}), incorporating heuristics, or using more data, while helpful, might not be enough to fully solve the object detection task. As an example, detecting mirrors or windows in images demands high-level reasoning and scene understanding. Please see~\cite{yang2019mirror} and Fig.~\ref{fig:mirror}.

\noindent {\bf Acknowledgment.} I would like to thank Seyed Mehdi Iranmanesh (seiranmanesh@mix.wvu.edu) and Chufeng Tang (chufeng.t@foxmail.com) for their help in preparing results in section~\ref{sec:error} 6 and Table~\ref{tab:detdismiss}, respectively.

\bibliographystyle{spbasic}
\bibliography{egbib}

\end{document}